\documentclass[sigconf]{acmart}

\AtBeginDocument{%
  \providecommand\BibTeX{{%
    \normalfont B\kern-0.5em{\scshape i\kern-0.25em b}\kern-0.8em\TeX}}}

\usepackage{algorithm}
\usepackage{algorithmic}
\usepackage{pdfpages}
\usepackage{multirow}
\usepackage[export]{adjustbox} 
\usepackage{bm}
\usepackage{amsmath}
\usepackage{array}
\usepackage{booktabs}
\usepackage{footmisc}

\usepackage{amsthm}
\usepackage{graphicx}
\usepackage{enumitem}

\theoremstyle{definition}
\newtheorem{definition}{Definition}
\usepackage{subfigure}
\usepackage{hyperref}
\usepackage{url}
\usepackage{color}
\usepackage{xcolor}
\usepackage{colortbl}
\usepackage{xspace}
\usepackage{verbatim}

\newcommand{\model}{CorGCN\xspace}

\usepackage{amsmath,amsfonts,bm}

\def\eqref#1{equation~\ref{#1}}

\def\1{\bm{1}}

\def\mC{{\bm{C}}}

\DeclareMathAlphabet{\mathsfit}{\encodingdefault}{\sfdefault}{m}{sl}
\SetMathAlphabet{\mathsfit}{bold}{\encodingdefault}{\sfdefault}{bx}{n}

\def\gE{{\mathcal{E}}}

\def\gG{{\mathcal{G}}}

\def\gL{{\mathcal{L}}}

\def\gN{{\mathcal{N}}}

\def\gV{{\mathcal{V}}}

\copyrightyear{2025}
\acmYear{2025}
\setcopyright{acmlicensed}\acmConference[KDD '25]{Proceedings of the 31st ACM SIGKDD Conference on Knowledge Discovery and Data Mining V.1}{August 3--7, 2025}{Toronto, ON, Canada}
\acmBooktitle{Proceedings of the 31st ACM SIGKDD Conference on Knowledge Discovery and Data Mining V.1 (KDD '25), August 3--7, 2025, Toronto, ON, Canada}
\acmDOI{10.1145/3690624.3709197}
\acmISBN{979-8-4007-1245-6/25/08}

\begin{document}

\title{Correlation-Aware Graph Convolutional Networks for Multi-Label Node Classification}

\author{Yuanchen Bei}
\affiliation{%
  \institution{Zhejiang University}
  \city{Hangzhou}
  \country{China}}
\email{yuanchenbei@zju.edu.cn}

\author{Weizhi Chen}
\affiliation{%
  \institution{Zhejiang University}
  \city{Hangzhou}
  \country{China}
}
\email{chenweizhi@zju.edu.cn}

\author{Hao Chen}
\affiliation{%
  \institution{The Hong Kong Polytechnic University}
  \city{HongKong SAR}
  \country{China}
}
\email{sundaychenhao@gmail.com}

\author{Sheng Zhou}
\authornote{Corresponding author.}
\affiliation{%
  \institution{Zhejiang University}
  \city{Hangzhou}
  \country{China}
}
\email{zhousheng\_zju@zju.edu.cn}

\author{Carl Yang}
\affiliation{%
  \institution{Emory University}
  \city{Atlanta}
  \country{USA}
}
\email{j.carlyang@emory.edu}

\author{Jiapei Fan}
\affiliation{%
  \institution{Alibaba Group}
   \city{Hangzhou}
  \country{China}
}
\email{jiapei.fjp@@alibaba-inc.com}

\author{Longtao Huang}
\affiliation{%
  \institution{Alibaba Group}
   \city{Hangzhou}
  \country{China}
}
\email{kaiyang.hlt@@alibaba-inc.com}

\author{Jiajun Bu}
\affiliation{%
  \institution{Zhejiang University}
  \city{Hangzhou}
  \country{China}
}
\email{bjj@zju.edu.cn}

\renewcommand{\shortauthors}{Yuanchen Bei et al.}

\begin{abstract}
Multi-label node classification is an important yet under-explored domain in graph mining as many real-world nodes belong to multiple categories rather than just a single one. 
Although a few efforts have been made by utilizing Graph Convolution Networks (GCNs) to learn node representations and model correlations between multiple labels in the embedding space, they still suffer from the ambiguous feature and ambiguous topology induced by multiple labels, which reduces the credibility of the messages delivered in graphs and overlooks the label correlations on graph data.
Therefore, it is crucial to reduce the ambiguity and empower the GCNs for accurate classification.
However, this is quite challenging due to the requirement of retaining the distinctiveness of each label while fully harnessing the correlation between labels simultaneously. 
To address these issues, in this paper, we propose a \textbf{Cor}relation-aware \textbf{G}raph \textbf{C}onvolutional \textbf{N}etwork (\textbf{\model}) for multi-label node classification. 
By introducing a novel Correlation-Aware Graph Decomposition module, CorGCN can learn a graph that contains rich label-correlated information for each label. It then employs a Correlation-Enhanced Graph Convolution to model the relationships between labels during message passing to further bolster the classification process. 
Extensive experiments on five datasets demonstrate the effectiveness of our proposed CorGCN.

\end{abstract}




\maketitle

\section{Introduction}
Node classification serves as a cornerstone in the field of graph mining~\cite{xiao2022graph,cai2018comprehensive}. 
Over the past decade, Graph Convolutional Networks (GCNs) have achieved remarkable success in this area by aggregating information from neighboring nodes, where edges in the graph often indicate similarities in the single-label space~\cite{kipf2017gcn,velivckovic2018gat}. 
However, graphs in real-world scenarios often entail multi-label nodes.
For example, users in social networks usually exhibit broad interests and embody multiple labels~\cite{zhang2018broad,sun2018attentive}, and protein nodes in protein interaction networks usually carry several relevant gene ontology annotations~\cite{zeng2019graphsaint,gao2023hierarchical}.
The diverse labels offer varied perspectives for delineating node characteristics while also introducing both challenges and opportunities for GCNs in the realm of multi-label node classification.

\begin{figure}
\centering
\subfigure[Toy Example of Multi-Label Node Classification]{
    \centering
    \includegraphics[width=0.95\linewidth]{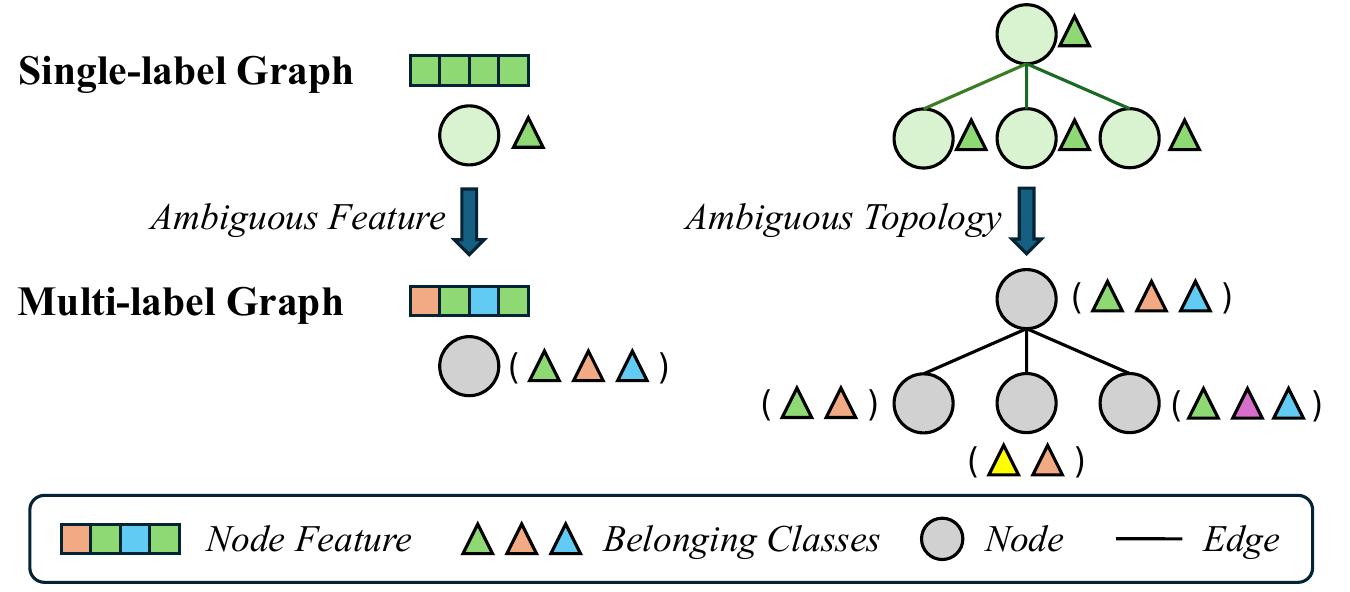}
  }
  \centering
  \subfigure[Ambiguous Feature]{
    \centering
    \includegraphics[width=0.476\linewidth]{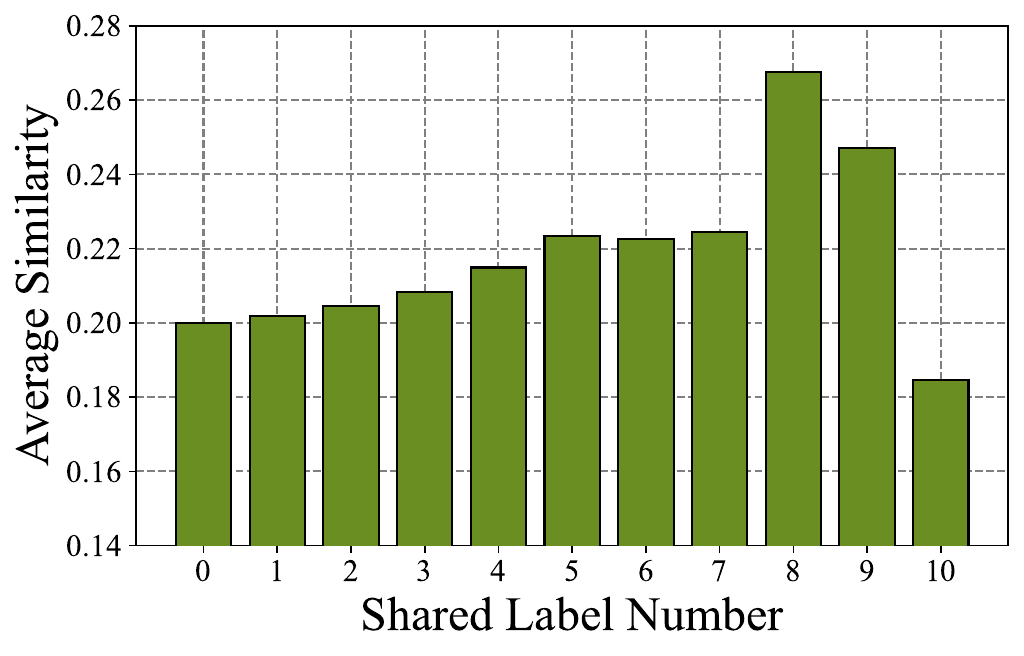}
  }
  \subfigure[Ambiguous Topology]{
    \centering
    \includegraphics[width=0.476\linewidth]{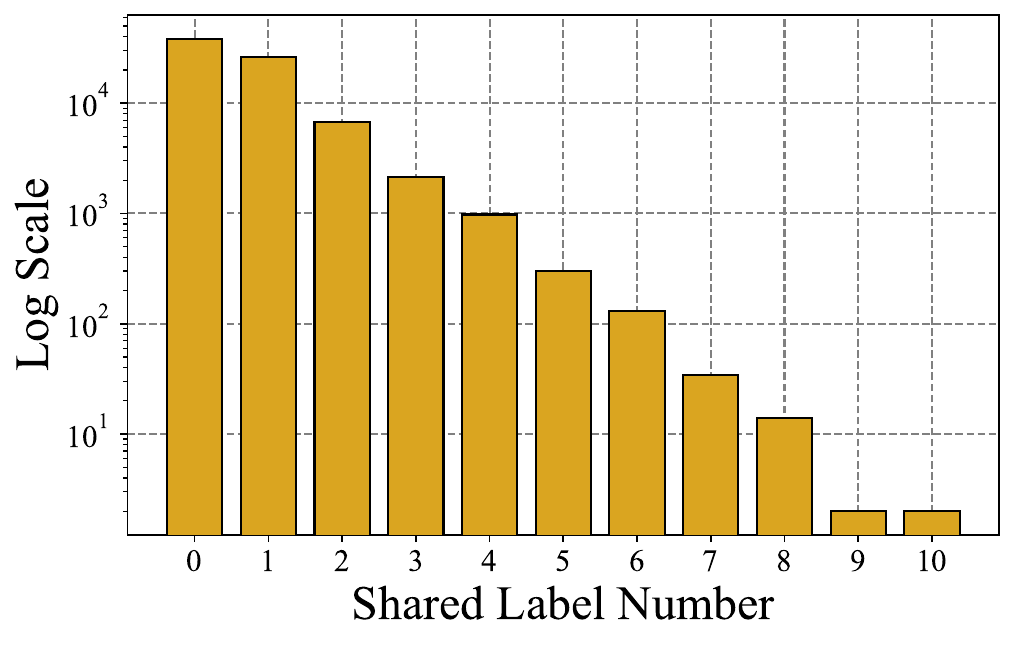}
  }
  \vspace{-0.8em}
  \caption{(a) A toy example of multi-label node classification challenges. (b)-(c) Illustrations of the ambiguous feature and ambiguous topology phenomenons on the PCG dataset.}
  \vspace{-0.8em}
  \label{fig:intro}
\end{figure}

Inspired by the success of multi-label learning on non-relational data~\cite{liu2017deep,wu2020multi,wang2020multi}, such as images and text, there has been a growing interest in applying multi-label learning to relational data, including graphs~\cite{gao2019mlgcn,xiao2022larn,zhou2021lanc}. 
Among the few existing methods for multi-label node classification, most employ GCNs to map nodes to low-dimensional representations. They then follow the traditional multi-label learning paradigm to model the relationships between nodes and labels, as well as the inter-label relationships in the representation space. However, this paradigm overlooks the characteristics of graph data in multi-label scenarios, as Figure \ref{fig:intro}-(a), leading to challenges for GCNs.
(1) \textbf{Ambiguous Feature}: In the single-label setting, nodes can aggregate patterns specific to a particular type of label from one neighborhood node by transforming the features.
However, in the multi-label setting, the feature of one node may be related to multiple labels and the patterns represented by the feature are ambiguous.
As a result, aggregating information from such ambiguous features will significantly impair the discriminative ability of the representation. 
Figure \ref{fig:intro}-(b) illustrates that nodes that allocate similar features may share a diverse number of labels in real-world graphs. 
(2) \textbf{Ambiguous Topology}:
In the single-label setting, connected nodes typically share the same single label (also known as the edge homophily assumption), so the pattern propagated along the edge is usually deterministic. 
However, in the multi-label setting, both the connected nodes have multiple labels, and the patterns propagated along the edge are often ambiguous. 
This ambiguity makes it difficult to determine from which connected nodes we should aggregate specific label information. 
Aggregating information directly from all neighbors would further accumulate ambiguity, compromising the discriminative power of the representations learned by GCNs and eventually impacting the inferring of specific labels.
Figure \ref{fig:intro}-(c) illustrates that edge-connected nodes may share diverse numbers of labels in real-world graphs.
\textit{It is essential to reduce the ambiguity so that the potential of GCNs in multi-label node classification can be fully released.}

Although important, addressing the above issues is non-trivial and meets the following challenges:
(1) \textbf{Label Distinctiveness}: As previously discussed, node attributes and edges in a multi-label graph may be influenced by multiple labels simultaneously. Directly extracting information from such a graph with a mixture of node labels can result in the loss of label distinctiveness and lead to inadequate exploration of each label.
(2) \textbf{Label Correlation}: In the multi-label setting, a node's association with multiple labels implies a correlation between these labels. The success of existing multi-label learning methods also demonstrates that fully leveraging these inter-label correlations can significantly enhance the quality of the representation~\cite{tarekegn2024deep,zhao2021hot}. Consequently, merely extracting messages or neighbors for a single label only could forfeit these vital correlations, leading to suboptimal outcomes.
This raises a pertinent question: \textit{How can we retain the distinctiveness of each label while fully harnessing the correlation between labels to achieve more accurate multi-label learning?}

To address the aforementioned challenges, in this paper, we present (\textbf{\model}), a \textbf{\underline{Cor}}relation-Aware \textbf{\underline{G}}raph \textbf{\underline{C}}onvolutional \textbf{\underline{N}}etwork for multi-label node classification. To tackle the first challenge, we propose a novel graph decomposition strategy where we learn an individual graph for each label while preserving its unique characteristics.
To tackle the second challenge, we fully consider the correlation between labels during the graph decomposition process. Instead of indiscriminately discarding all information from other categories, we retain the information of related labels within each category's graph.
Lastly, based on the multiple graphs generated, we further exploit the correlation between labels using a newly designed correlation-enhanced GCN that takes category correlation into account, thereby enhancing the final multi-label classification results.
We conduct extensive experiments on five datasets with comprehensive metrics. The results demonstrate the effectiveness of the proposed method. The primary contributions of this paper can be summarized as follows:

\begin{itemize}[leftmargin=*]
    \item We highlight the overlooked impact of ambiguous feature and ambiguous topology on GCNs in the multi-label node classification, which is fundamental in this critical task.
    \item We propose a Correlation-aware Graph Convolutional Network (CorGCN) to simultaneously retain label-distinct characters and label correlation information for multi-label node classification. 
    \item Extensive experiments on five datasets demonstrate \model significantly outperforms nine state-of-the-art baselines across seven metrics. Further in-depth analysis from diverse perspectives also demonstrates the strengths of \model.
\end{itemize}

\vspace{-0.5em}

\section{Related Works}
\subsection{Multi-Label Learning}
Multi-label learning, where each instance is associated with a set of correlated labels rather than a single label, is widely applicable in various domains such as object detection~\cite{ge2018odmll} and text classification~\cite{xiao2019tcmll}.
For example, an image may contain multiple 
relevant objects and a text can encompass various topics~\cite{liu2021emerging,bogatinovski2022survey,liu2017deep,wu2020multi}.
The key to multi-label learning is to model the multi-label correlation. Early studies convert the multi-label learning problem into multiple single-label learning problems, which has limitations in its ability to explicitly model correlations~\cite{boutell2004br,read2009cc}. Recently, various studies have been proposed with deep representation learning. Among them, one type of method learns correlated representations with statistical analysis~\cite{yeh2017learning,wang2018deepcca}. The second type of method learns the label correlation on label embeddings with label sequence/relational modeling techniques, such as RNN and GNN~\cite{wang2016cnnrnn,yazici2020orderless,durand2019learninggnn}.
Another type of model proposes to co-learn label and instance representations with the auto-encoder architecture, which models the correlation during the latent encoding~\cite{bai2021disentangled,bai2022gaussian,zhao2021hot}.

Despite significant progress, these methods are all focused on Euclidean data. In recent years, due to the development of GNNs, the problem of multi-label learning on non-Euclidean graphs has also shown its importance~\cite{zhao2023multi}. However, thus far, multi-label learning on graph data has not received much attention.

\subsection{Multi-Label Node Classification}
Node classification is a well-known fundamental task in graph data mining. Typically, node classification refers to the assignment of a unique label to each node~\cite{grover2016node2vec,bhagat2011node}. In recent years, multi-label node classification (MLNC) has posed significance due to the increasing realization that nodes on graphs often exhibit multiple associated categories simultaneously~\cite{zhao2023multi,gao2019mlgcn}. For example, in social network analysis, a user often belongs to multiple interest groups~\cite{song2021semi,wu2022graph,chen2024macro}. This involves assigning non-unique and variable numbers of labels to each node.
Currently, MLNC is still in its infancy, and only a few studies have focused on MLNC~\cite{zhao2023multi,song2021semi}. Representatively, ML-GCN~\cite{gao2019mlgcn} learns distinct embeddings for each label and constructs node-label correlations to augment the GCN-encoded node embeddings. Subsequently, LANC~\cite{zhou2021lanc} and LARN~\cite{xiao2022larn} further incorporate attention mechanisms to integrate label embeddings with GCN-encoded node embeddings more effectively. 

Nonetheless, current approaches continue to utilize a unified graph convolution process across neighborhoods under various labels, which ignores the possibility that information among different multi-label nodes may become intermixed in a unified message passing process, including both correlated and uncorrelated labels.

\subsection{Graph Structure Learning}
Since the natural graph structure often contains noise information, it's not always directly applicable to a wide range of downstream tasks. This limitation has brought the field of Graph Structure Learning (GSL) into the spotlight, garnering increasing attention in recent years~\cite{li2024gslb,bei2023reinforcement}. Representatively, GRCN~\cite{yu2021grcn} adjusts edge weights in the graph, optimizing with downstream tasks. IDGL~\cite{chen2020idgl} iteratively improves the graph structure through node embeddings. And SUBLIME~\cite{liu2022sublime} uses a contrastive-based objective function to guide the learned graph.
These GSL models generally strive to refine the graph structure by enhancing graph homophily or by augmenting task-specific useful information in the single-label space. Despite these advances, there is a noticeable absence of GSL approaches specifically tailored for MLNC, leaving a significant gap in the field.

In light of these challenges, our paper is dedicated to introducing a GSL-based methodology, uniquely designed for MLNC. This method aims to address the existing limitations by integrating the understanding of multi-label correlations and adapting the graph structure accordingly.

\section{Preliminaries}

\textbf{Notations}. 
Given a graph $\gG = (\gV, \gE)$, consisting of two sets: a set of nodes $\gV$ and a set of edges $\gE$. Here, $\gV$ represents the set of nodes in the graph, while $\gE \subseteq \{(u, v) | u, v \in \gV\}$ represents the set of edges between the nodes.
Then, the adjacency matrix of $\gG$ can be defined as $\bm{A}\in \mathbb{R}^{n\times n}$ with $m$ non-zero element in the matrix based on the edge set, where $n=|\gV|$ and $m=|\gE|$ are the node number and edge number, respectively.
Each node $v_i\in \gV$ in the graph $\gG$ is associated with a feature vector $\bm{x}_{i} \in \mathbb{R}^{f}$ and thus all the node feature vectors form the overall feature matrix $\bm{X}\in \mathbb{R}^{n\times f}$ of the graph $\gG$. To distinguish from the notation of decomposed graphs introduced in this paper, we represent the original graph adjacent matrix with the symbol $\bm{A}^{0}$.

\begin{definition}[\textit{Graph with Multi-Label Nodes}]
In a multi-label graph, each node \( v_i \in \gV \) is associated with a set of labels \( L_i \subseteq L \), where \( L \) is the set of all possible labels and $|L| = K$ is the total number of labels. Thus, each node can have multiple labels, i.e., \( |L_i| \ge 1 \).
In practice, each node $v_i$ will be assigned a multi-hot label vector $\bm{y}_{i}\in \mathbb{R}^{1\times K}$, where each element $y_{i,j} \in [0, 1]$ in $\bm{y}_{i}$ indicates whether node $v_i$ belongs to class $j$.
\end{definition}

\begin{definition}[\textit{Multi-Label Node Classification}]
In the multi-label node classification scenario, we typically have a subset of nodes with known labels and another subset with unknown labels~\cite{zhao2023multi}. 
The node set \( \gV \) in $\gG$ can be divided into two disjoint subsets: a labeled node set \( \gV_L \) and an unlabeled node set \( \gV_U \). Thus, \( \gV = \gV_L \cup \gV_U \) and \( \gV_L \cap \gV_U = \emptyset \).
The goal of multi-label node classification is to learn a model \( f: \gV \rightarrow \{0,1\}^{K} \) using the labeled nodes \( v_i \in \gV_L \) and the graph topology $\bm{A}$. This model aims to predict the label set \( \bm{y}_j \) for unlabeled nodes \( v_j \in \gV_U \).
\end{definition}

\section{Methodology}
\begin{figure*}[!t]
     \centering
     \includegraphics[width=\linewidth, trim=0cm 0cm 0cm 0cm,clip]{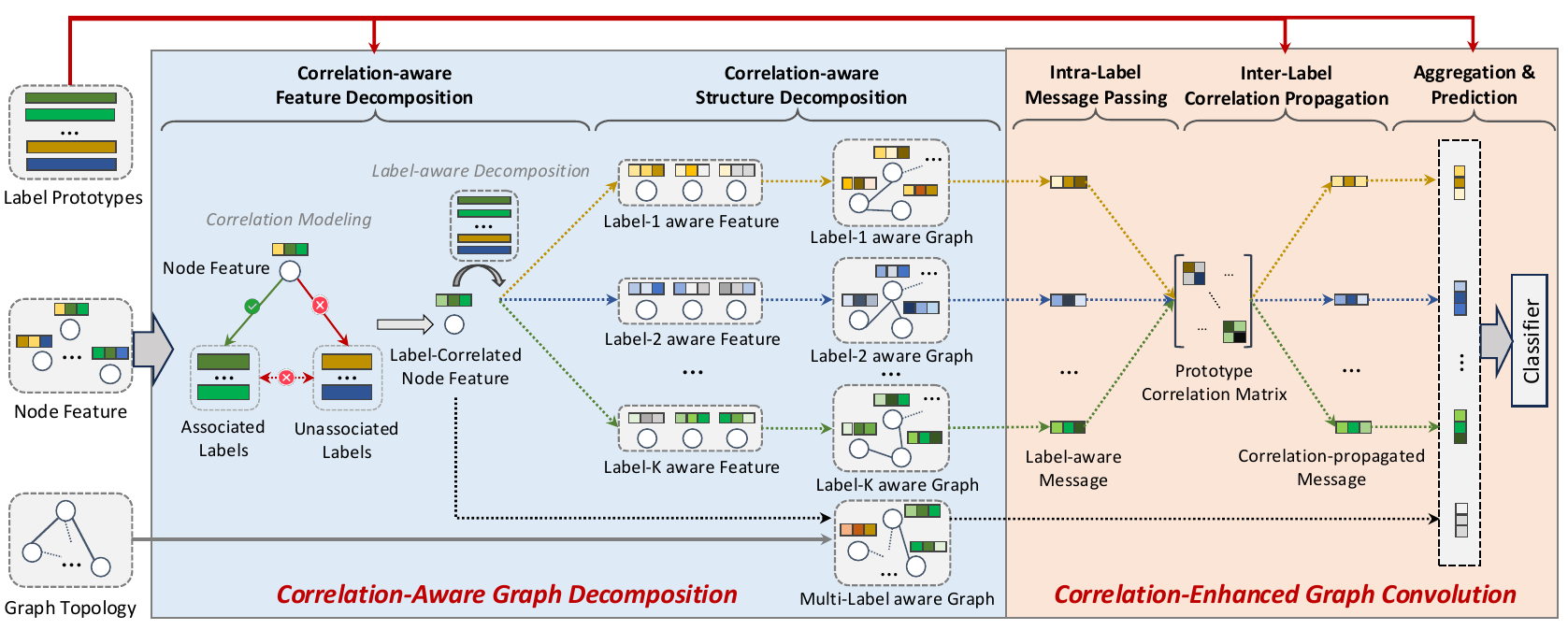}
     \vspace{-0.5em}
     \caption{The overall architecture of \model.
     (a) Correlation-Aware Graph Decomposition: it first learns label-correlated node features and decomposes them into multiple label-aware features. Then, based on the decomposed features, it decomposes multiple label-aware graphs.
     (b) Correlation-Enhanced Graph Convolution: each layer includes intra-label message passing of the neighborhood in each label-aware graph view and inter-label correlation propagation between label-aware messages.
     }
     \vspace{-0.5em}
     \label{fig:framework}
\end{figure*}

Our proposed model for multi-label node classification comprises the correlation-aware graph learning module with node feature and graph topology decomposition (Sec.~\ref{sec:comp1}) and the correlation-aware graph convolution (Sec.~\ref{sec:comp2} and Sec.~\ref{sec:comp3}). Further, we illustrate how \model can be expanded to the scenario with large label space (Sec.~\ref{sec:extend_macro}). 
Figure~\ref{fig:framework} provides an overview of the architecture of our proposed \model.

\subsection{Correlation-Aware Graph Decomposition}\label{sec:comp1}
Due to the ambiguity in node features and topological structures of graphs in multi-label scenarios, we aim to decompose them into multiple graphs. However, directly performing the decomposition would result in the loss of the critically important multi-label correlation property. Therefore, we need to carry out the decomposition based on representations that have encapsulated label correlation.

\subsubsection{\textbf{Correlation-Aware Feature Decomposition}}
This component primarily adopts an approach that first models the node-label correlation and label-label correlation, and then, based on this, performs the decomposition of node features across different label spaces. Summarizing these, we model the correlations with contrastive mutual information estimator $\phi$ and likelihood maximizing decoder $\theta$, which can be defined as,
\begin{equation}
    \mathop{\arg\max}\limits_{\phi, \theta}\underbrace{I_{\phi}(\bm{E}^{x}, \bm{E}^{l})}_{\text{Mutual Information}} + \underbrace{\mathcal{L}_{\theta}(\theta; \bm{E}^{x}) + \mathcal{L}_{\theta}(\theta; \bm{E}^{l})}_{\text{Likelihood}}.
\end{equation}

Specifically, to fully utilize the knowledge from labeled nodes for generalization to all nodes in the graph, we first assign a trainable prototype embedding to each label to describe the characteristics of this label:
\begin{equation}
    \bm{E}^{l} = [\bm{E}^{l}_{1}, \cdots, \bm{E}^{l}_{k}, \cdots, \bm{E}^{l}_{K}],
\end{equation}
where $1\leq k \leq K$, and $\bm{E}^{l}_{k}\in \mathbb{R}^{d}$ is the $k$-th label prototype. 
We then transform the original node features $\bm{X}$ to the same $d$-dimensional space using the linear transformation: $\bm{E}^{x} = \bm{X}\bm{W}_t$, where $\bm{W}_t\in \mathbb{R}^{f\times d}$ is a learnable transformation matrix, and each vector $\bm{E}^{x}_{i} \in \mathbb{R}^{d}$ in $\bm{E}^{x}$ denotes the transformed feature of node $v_i$:
\begin{equation}
    \bm{E}^{x} = [\bm{E}^{x}_{1}, \cdots, \bm{E}^{x}_{i}, \cdots, \bm{E}^{x}_{n}].
\end{equation}

Then, the mutual information estimator $\phi$ captures both node-label and label-label correlations based on those supervised nodes.
Previous works for multi-label learning have successfully adopted contrastive learning to model multi-label correlations and inspired by~\cite{bai2022gaussian}, we utilize contrastive learning as $\phi$ here.
Given the labeled nodes $\gV_{L}$, the contrastive-based mutual information estimator with $\mathcal{L}_{cmi}$ can be written as follows:
\begin{equation}\label{eq:cons}
    \gL_{cmi} = -\frac{1}{|\gV_{L}|}\sum_{i \in \gV_{L}}\frac{1}{|\text{Pos}(\bm{y}_{i})|}\sum_{p\in \text{Pos}(\bm{y}_{i})}\text{log}\frac{\text{exp}(\bm{E}^{x}_{i} \cdot \bm{E}^{l}_{p})}{\sum_{t\in \bm{y}_{i}}\text{exp}(\bm{E}^{x}_{i} \cdot \bm{E}^{l}_{t})},
\end{equation}
where $\text{Pos}(\bm{y}_{i})$ denotes the label set that node $v_i$ belongs to. 
Through this formulation, we directly map node features into a compact space that reflects multi-label correlations, while implicitly capturing the nuances of multi-label prototype relationships. 
Unlike single-label scenarios where classes are mutually exclusive, multi-label classification acknowledges the complexity of label correlations.
Therefore, we do not enforce strict contrastive relations among labels, maintaining their intrinsic associations~\cite{bai2022gaussian,zhao2021hot}. Instead, it utilizes feature embeddings as anchors and label embeddings as positive or negative examples, allowing frequently co-occurring labels to become more similar over time.

Subsequently, due to the above process mainly focuses on embedding space without explicit guidance, in order to ensure that different node features and label prototypes retain their own classification characteristics, we further constrain this through likelihood maximizing decoder with corresponding loss $\mathcal{L}_{lm}$ as follows:

\begin{equation}
\rho_{k}= \frac {\sqrt{\frac{1}{\sum_{v_i \in \gV_L}\bm{y}_{i,k}}   }}   {\sum^{K}_{j=1}  \sqrt{\frac{1}{\sum_{v_i \in \gV_L}\bm{y}_{i,j}}} },
\end{equation}

\begin{equation}\label{eq:le}
\begin{aligned}
    \mathcal{L}_{lm} = -\frac{1}{2|\gV_L|} \sum_{v_i \in \gV_L} \sum_{k=1}^{K} &\rho_{k}[(1 - p_{k}(\bm{E}_{i}^x | \theta))^{\gamma}\cdot \text{log}p_{k}(\bm{E}_{i}^x | \theta) \\
    &+ (1 - p_{k}(\bm{y}_{i}\cdot \bm{E}^{l} | \theta))^{\gamma}\cdot\text{log}p_{k}(\bm{y}_{i}\cdot \bm{E}^{l} | \theta)],
\end{aligned}
\end{equation}
where $\rho_{k}$ is a statistical parameter to enhance the learning for classes with a small number of node samples and $\gamma$ is a hyper-parameter to control the observation of hard node/label samples~\cite{lin2017focal}. $\hat{p}_{k}(\bm{E}_{i}^x | \theta)$ is the predicted result of the decoder $\theta$, and if $y_{i,k} = 1$, $p_{k}(\bm{E}_{i}^x | \theta) = \hat{p}_{k}(\bm{E}_{i}^x | \theta)$. Otherwise, $p_{k}(\bm{E}_{i}^x | \theta) = 1 - \hat{p}_{k}(\bm{E}_{i}^x | \theta)$.

In the above process, we obtain the correlated node features $\bm{E}^{x}$ and label prototypes $\bm{E}^{l}$. Each label prototype is associated with a specific label and its potentially correlated labels while each correlated node feature is still ambitious due to the mixture of multiple label information.
Based on this, we decompose the correlated node features to each label-friendly feature space that \textit{considers both the label-specific and its correlated label-specific information} with the guidance of label prototypes.
Specifically, given the label-correlated node feature $\bm{E}^{x}_{i}$ of node $v_i$ and the label prototype of label $k$, the projection process can be obtained by:
\begin{equation}
    w_{i,k} = \text{sim}(\bm{E}^{x}_{i}, \bm{E}^{l}_{k}) =  \frac{\bm{E}^{x}_{i}\cdot\bm{E}^{l}_{k}}{\|\bm{E}^{x}_{i}\| \|\bm{E}^{l}_{k}\|},
\end{equation}
\begin{equation}
\begin{aligned}
    \bm{E}^{proj}_{i} &= [\bm{E}^{proj}_{i, 1}, \cdots, \bm{E}^{proj}_{i, k}, \cdots, \bm{E}^{proj}_{i, K}] \\
    &= [w_{i,1}\bm{E}^{x}_{i}, \cdots, w_{i,k}\bm{E}^{x}_{i}, \cdots, w_{i,K}\bm{E}^{x}_{i}],
\end{aligned}
\end{equation}
where $w_{i,k}$ is the projection coefficient of node $i$ to label $k$, $\bm{E}^{proj}_{i,k}$ is the projected representation of node $v_i$ towards the $k$-th label.

\subsubsection{\textbf{Correlation-Aware Structure Decomposition}}
Based on the correlation-aware decomposed node features $\bm{E}^{proj}$, we aim to decompose the graph structure (message passing path) for each label with its correlated labels.

To preserve the graph topology patterns, we first aggregate the neighborhood projected features of each center node $v_i$ in each label view $k$ for structure decomposition:
\begin{equation}
    \bm{E}^{sd}_{i,k} = \text{Agg}(\bm{E}^{proj}, \bm{A}^{0}) 
     = \frac{1}{|\gN^{0}_{i}|+1}\sum_{j\in\{\gN^{0}_{i}\cup \{v_{i}\}\}}\bm{E}^{proj}_{j,k},
\end{equation}
where $\gN^{0}_{i}$ is the neighborhood of node $v_i$ in original graph adjacent matrix $\bm{A}^{0}$.
We leverage the aggregated representations to measure the node similarity in each label view:
\begin{equation}
    S^{k}_{i,j} = \text{sim}(\bm{E}^{sd}_{i,k}, \bm{E}^{sd}_{j,k}) = \frac{\bm{E}^{sd}_{i,k}\cdot\bm{E}^{sd}_{j,k}}{\|\bm{E}^{sd}_{i,k}\| \|\bm{E}^{sd}_{j,k}\|},
\end{equation}
where $\bm{S}^{k}$ is the similarity score matrix in label view-$k$, $S^{k}_{i,j}$ is the score between node $v_i$ and node $v_j$.

Then, for each node $v_i$, the structure decomposing for different label-aware graphs can be described as:
\begin{equation}
A^{k}_{i,j} = 
\begin{cases} 
    1, &  S^{k}_{i,j}\in \text{Top-$\lambda$}(\bm{S}^{k}_{i}); \\
    0, & \text{\textit{otherwise}}.
\end{cases}
\end{equation}
\begin{equation}
    \gG^{k} = (\bm{A}^{k}, \bm{E}^{proj}_{k}),
\end{equation}
where $A^{k}$ is the adjacent matrix of the $k$-th label view, and $\lambda$ is the hyperparameter that controls the density of the decomposed graph.

Furthermore, with the correlated label-aware node feature and the original graph topology, the multi-label aware graph $\gG^{0} = (\bm{A}^{0}, \bm{E}^{x})$ can be obtained to capture correlated structure patterns by message passing, and thus the learned correlation-aware decomposed graphs $CDG$ can be obtained as follows:
\begin{equation}
    CDG = \{\gG^{0}, \gG^{1}, \cdots, \gG^{K}\}.
\end{equation}

\subsection{Correlation-Enhanced Graph Convolution}\label{sec:comp2}
Previous studies for multi-label node classification mainly conduct unified neighborhood message passing~\cite{gao2019mlgcn,xiao2022larn,zhou2021lanc}.
We contend this approach has two primary limitations: (1) passing ambiguous messages from the neighborhood in a unified manner; and (2) correlation ignorance, which overlooks the label correlation when passing the message.
Therefore, we further equip the $CDG$ with Correlation-Enhanced graph convolution.

\subsubsection{\textbf{Intra-Label Message Passing}}
Firstly, Correlation-Enhanced graph convolution conducts intra-label message passing within the graph $\gG^{k}$ of each label view $k$:
\begin{equation}
\begin{aligned}
    \bm{\hat{Z}}^{(l)}_{[:, k]} & = \text{GCN}(\bm{Z}^{(l-1)}_{[:, k]}, \tilde{\bm{A}}^{k})\\ &=   \sigma(\tilde{\bm{A}}^{k}\bm{Z}^{(l-1)}_{[:, k]}\bm{W}^{(l)}),
\end{aligned}\label{eq:gnn_mp}
\end{equation}
where $0\leq k \leq K$, $\tilde{\bm{A}^{k}}=\hat{\bm{D}}^{k-\frac{1}{2}}\bm{A}^{*k}\hat{\bm{D}}^{k-\frac{1}{2}} \in \mathbb{R}^{n \times n}$ is the normalized adjacency matrix of label view-$k$, $\hat{\bm{D}^{k}} \in \mathbb{R}^{n \times n}$ is the degree matrix of $\bm{A}^{*k}=\bm{A}^{k}+\bm{I}$ where $\bm{I}$ is the identity matrix. $\bm{\hat{Z}}^{(l)}$, $\bm{W}^{(l)}$ is the output features, trainable parameters in $l$-th message passing layer, respectively. The input of the first layer $\bm{Z}^{(0)}$ = $[\bm{E}^{x}, \bm{E}_{1}^{proj}, ..., \bm{E}_{K}^{proj}]\in \mathbb{R}^{n\times (K+1)\times d}$ is the feature matrix. Note that this is a GCN-like~\cite{kipf2017gcn} message passing function, any other graph message passing functions~\cite{velivckovic2018gat,NIPS2017_graphsage} can also be adopted here.

\subsubsection{\textbf{Inter-Label Correlation Propagation}}
After the above intra-label message passing, we obtain the neighborhood messages $\bm{\hat{Z}}^{(l)}\in \mathbb{R}^{n\times K \times d}$ from each label-aware view. Then, to model their correlations, we further propose the inter-label correlation propagation between each label-aware graph view:

\begin{equation}
    \bm{Cor}_{i} = \text{Softmax}(\frac{(\bm{E}^l \bm{W}_1)(\bm{\hat{Z}}^{(l)}_{[i,1:]} \bm{W}_2)^T}{\sqrt{d}}),
\end{equation}
\begin{equation}
    \bm{Z}^{(l)}_{[i,1:]} = \bm{Cor}_{i} \bm{\hat{Z}}^{(l)}_{[i,1:]} \bm{W}_3,
\end{equation}
where $\bm{Cor}_{i} \in \mathbb{R}^{K\times K}$ is the label prototype correlation matrix for node $v_i$, $\bm{W}_1$, $\bm{W}_2$, $\bm{W}_3$ are the trainable parameters, and $\bm{Z}^{(l)}_{[i,1:]} \in \mathbb{R}^{K\times d}$ is the output of $v_i$ in $l$-th graph convolution layer containing $K$ views of inter-label correlated representations.

\subsubsection{\textbf{Aggregation \& Prediction}}
After the multi-layer graph message passes, we aggregate these representations to obtain the final representation with both the node representations and label prototypes. The aggregation process can be written as:
\begin{equation}
    \bm{Z}^{cls}_i = [\bm{Z}_{i,0} || (\sum_{k=1}^{K} \text{sim}(\bm{Z}_{i,k}, \bm{E}^{l}_{k})\cdot \bm{Z}_{i,k})],
\end{equation}
\begin{equation}
    \hat{\bm{y}}_{i} = \sigma(\bm{W}^{cls}\bm{Z}^{cls}_{i}+\bm{b}^{cls}),
\end{equation}
where $\bm{Z}_{i,k}$ is output features from final layers of correlation propagation for node $v_i$ in the label view $k$, $\bm{Z}^{cls}_i \in \mathbb{R}^{n\times 2d}$ is the final node representation for multi-label node classification of node $v_i$ and $||$ is the concatenation operation, $\bm{W}^{cls}$ and $\bm{b}^{cls}$ are the trainable parameters in the classifier, and $\sigma$ is the Sigmoid function.

\subsection{Objective Function}\label{sec:comp3}
To train and optimize the model parameters, we apply the binary cross-entropy loss as the model objective function.
Formally, for each node $v_i \in \gV_L$, the classification objective function can be expressed as:
\begin{equation}
    \gL_{cls} = -\frac{1}{|\gV_L|} \sum_{i\in \gV_L} \frac{1}{K}\sum_{j=1}^{K}y_{i,j} \log(\hat{y}_{i,j}) + (1-y_{i,j}) \log(1-\hat{y}_{i,j}),
\end{equation}
where $\hat{y}_{i,j}$ is the predicted logits of node $v_i$ towards label $j$ and $y_{i,j}$ is the corresponding ground-truth label.
Then, considering the objective losses in each \model module, the overall objective function of \model is as follows:
\begin{equation}
    \gL=\gL_{cls} + \alpha \cdot \gL_{cmi} + \beta \cdot \gL_{lm},
\end{equation}
where $\alpha$ and $\beta$ are the controllable hyper-parameters. 

\subsection{Extension to Large Label Space}\label{sec:extend_macro}
In some real-world graph structures, nodes may exist in a very large multi-label space, such as in large protein interaction networks, where individual protein nodes may have hundreds of characteristic labels simultaneously~\cite{jing2021fast,rong2020self}.
In this subsection, we demonstrate that our proposed \model can be readily extended to multi-label node classification with large label space for efficient learning.

\textbf{Macro Label Prototypes.}
For the large label space, directly applying multi-label graph learning with $K$ prototypes will obtain hundreds of label-aware graphs, which is not efficient enough for applications. Therefore, we aim to refine and cluster the original prototypes into macro-label prototypes, with the principle that correlated labels will form a macro prototype. 

Specifically, we first pre-train the $K$ label prototypes in Eq.(1) with $\mathcal{L}_{cmi}$ and $\mathcal{L}_{lm}$ (as Sec. 4.1). Then, we adopt K-means-based clustering to refine the $K$ pre-trained label prototypes into $K'$ macro prototypes~\cite{hamerly2003learning}.
Given the pre-trained label prototypes $\bm{E}^{l}$, we randomly initialize $K'$ macro prototype centroids $\{\bm{\mu}_{1}, ..., \bm{\mu}_{k}, ..., \bm{\mu}_{K'}\}$, where $\bm{\mu}_{k}\in \mathbb{R}^{d}$ is the centroids of macro label prototype $\bm{C}_{k}$, $K' \ll K$ is the hyperparameter set as the macro prototype number.
Then, we explore and assign the original label prototypes to the appropriate macro label prototype based on their representation similarity, and update the centroid of the macro label prototype iteratively. The process can be expressed as:
\begin{equation}
    \bm{\mu}_{k} = \frac{1}{|\mC_{k}|}\sum_{x_{v}=k, \bm{E}^{l}_{v} \in \mC_{k}} \bm{E}^{l}_{v},
\end{equation}
where $|\mC_{k}|$ is the number of labels within $\mC_{k}$, and $x_{v}$ is the macro label prototype index that $v$ is assigned to. Further, the optimization target of the macro label prototype generation is:
\begin{equation}
\begin{aligned}
    \min_{x_1, ..., x_{K} \atop \bm{\mu}_{1}, .., \bm{\mu}_{K'}} &J\left(x_1, ..., x_{K}; \bm{\mu}_{1}, .., \bm{\mu}_{K'} \right)
    \\ \overset{\bigtriangleup}{=} & \sum_{k=1}^{K'}\sum_{x_{v}=k, \bm{E}^{l}_{v}\in \mC_{k}}\sqrt{(\bm{E}^{l}_{v}-\bm{\mu}_{k})(\bm{E}^{l}_{v}-\bm{\mu}_{k})^\top},
\end{aligned}
\end{equation}
where $J$ is the objective function of macro label prototype generation.
In this way, we obtain the macro label prototype set: 
\begin{equation}
    \bm{E}^{ml} = [\bm{\mu}_{1}, \cdots, \bm{\mu}_{k}, \cdots, \bm{\mu}_{K'}],
\end{equation}
where $1\leq k \leq K' \ll K$. Therefore, we can reduce the original number of label prototypes to meet our needs. With the obtained $\bm{E}^{ml}$, we can flexibly replace $\bm{E}^{l}$ with $\bm{E}^{ml}$ and procedure Sec.~\ref{sec:comp1} - Sec.~\ref{sec:comp3} in the same way. Model analyses are detailed in Appendix~\ref{sec:add_analysis}.

\section{Experiments}

In this section, we conduct comprehensive experiments on benchmark datasets with \model, aiming to answer the following questions:
\textbf{RQ1:} How does \model perform compared to representative and state-of-the-art models?
\textbf{RQ2:} What is the effect of different components in \model?
\textbf{RQ3:} Can \model be generalized and adapted to different GNN message passing backbones?
\textbf{RQ4:} How does the impact of \model on each fine-grained class?
\textbf{RQ5:} What is the efficiency of \model on training and inference?
\textbf{RQ6:} How do key hyper-parameters impact the performance of \model?

\subsection{Experimental Setup}
\subsubsection{Datasets}
We conduct experiments on five widely used benchmark datasets: \textbf{Humloc}~\cite{zhao2023multi}, \textbf{PCG}~\cite{zhao2023multi}, \textbf{Blogcatalog}~\cite{zhou2021lanc}, \textbf{PPI}~\cite{zeng2019graphsaint}, and \textbf{Delve}~\cite{xiao2022larn} to verify the effectiveness of \model. The statistics of these datasets are shown in Table \ref{tab:stats}. We randomly divided the labeled data of these datasets into training, validation, and test sets, following a 6:2:2 ratio, respectively~\cite{zhao2023multi}. The performance on PPI and Delve are conducted on the extension of \model, with 20 and 10 macro label prototypes, respectively. 
Detailed descriptions of these datasets are illustrated in Appendix \ref{sec:data_appendix}.

\begin{table}[htbp]
  \centering
  \small
  \caption{Statistics of the experimental datasets.}
  \resizebox{0.975\linewidth}{!}{
    \begin{tabular}{c|c|c|c|c|c}
    \toprule
    Dataset & \# Nodes & \# Edges & \# Features & \# Classes & Density \\
    \midrule
    Humloc & 3,106 & 18,496 & 32 & 14 & 0.3836\%\\
    PCG & 3,233 & 37,351 & 32 & 15 & 0.7149\% \\
    Blogcatalog & 10,312 & 333,983 & 100 & 39 & 0.6282\% \\
    PPI & 14,755 & 225,270 & 50 & 121 & 0.2070\% \\
    Delve & 1,229,280 & 4,322,275 & 300 & 20 & 0.0006\% \\
    \bottomrule
    \end{tabular}%
  }
  \label{tab:stats}%
\end{table}%

\subsubsection{Baselines}
To evaluate the effectiveness of \model, we compare it with nine representative and state-of-the-art models. 
(i) Traditional GNN for node classification: \textbf{GCN}~\cite{kipf2017gcn}. 
(ii) Graph structure learning methods: \textbf{GRCN}~\cite{yu2021grcn}, \textbf{IDGL}~\cite{chen2020idgl}, and \textbf{SUBLIME}~\cite{liu2022sublime}. 
(iii) Multi-label node classification methods: \textbf{GCN-LPA}~\cite{wang2021gcnlpa}, \textbf{ML-GCN}~\cite{gao2019mlgcn}, \textbf{LANC}~\cite{zhou2021lanc}, \textbf{SMLG}~\cite{song2021semi}, and \textbf{LARN}~\cite{xiao2022larn}.
Note that we equip all models with the GCN backbone for fair comparisons. Due to the traditional workflow of GCN being unsuitable for multi-label node classification, we replace the last activation function from softmax to sigmoid, with the activation value for the classification threshold set to 0.5 for each label.
The details of these baselines are left in Appendix \ref{sec:model_appendix}.

\subsubsection{Implementation Setting}
For all models, the embedding size is fixed to 64 for fair node classification.
The batch size is set to 1024 for all models and the Adam optimizer is used. Experiments are conducted on
the Centos system with NVIDIA RTX-3090 GPUs.
The detailed experiment implementations of \model are posed in Appendix~\ref{sec:imp_appendix} and
the source code of \model is available at the \href{https://github.com/YuanchenBei/CorGCN}{\textbf{Github}} link.

\subsubsection{Evaluation Metrics}
We evaluate the model under the setting for semi-supervised multi-label learning on graphs as previous works~\cite{song2021semi}.
We evaluate the models with seven widely-adopted metrics including \textit{Ranking Loss}, \textit{Hamming Loss}, \textit{Macro-AUC}, \textit{Micro-AUC}, \textit{Macro-AP}, \textit{Micro-AP}, and \textit{Label Ranking AP} (abbreviated as Ranking, Hamming, Ma-AUC, Mi-AUC, Ma-AP, Mi-AP, and LPAP in the following subsections, respectively).
All these multi-label node classification metrics are clearly defined in~\cite{zhang2013review}. 
Note that we run all the experiments \textit{five} times with different random seeds and report the average results with standard deviation.

\begin{table*}[htbp]
  \centering
  \caption{Multi-label node classification comparison results in percentage over \textit{five} trial runs ($\uparrow$: the higher, the better; $\downarrow$: the lower, the better). The best and second-best results in each column are highlighted in \textbf{bold} font and \underline{underlined}. OOM denotes out-of-memory during the model training. The average ranking is calculated based on the numerical result of each model.}
  \resizebox{\linewidth}{!}{
    \begin{tabular}{cc|ccccccccc|c}
    \toprule
    Dataset & Metrics & GCN   & GRCN  & IDGL  & SUBLIME & GCN-LPA & ML-GCN & LANC  & SMLG & LARN  & \textbf{\model} \\
    \midrule
    \multirow{7}[4]{*}{\rotatebox{90}{Humloc}} & Ranking (↓) & 13.29 ± 1.01 & 16.84 ± 0.09 & 19.04 ± 0.13 & 14.44 ± 0.10 & 19.58 ± 4.96 & \underline{13.28 ± 0.70} & 14.66 ± 0.36 & 17.82 ± 0.43 & 13.62 ± 0.46 & \textbf{12.57 ± 0.31} \\
          & Hamming (↓) & 7.52 ± 0.13 & 8.37 ± 0.03 & 8.40 ± 0.02 & 8.53 ± 0.03 & 8.30 ± 0.13 & \underline{7.46 ± 0.22} & 7.82 ± 0.16 & 8.55 ± 0.26 & 8.04 ± 0.29 & \textbf{7.37 ± 0.17} \\
\cmidrule{2-12}          & Ma-AUC (↑) & 69.10 ± 2.25 & 56.27 ± 0.23 & 50.26 ± 0.94 & 72.42 ± 0.40 & 57.91 ± 2.39 & 72.41 ± 2.47 & 68.73 ± 1.76 & \underline{72.64 ± 1.65} & 71.63 ± 3.30 & \textbf{77.31 ± 1.58} \\
          & Mi-AUC (↑) & 85.39 ± 1.30 & 82.85 ± 0.02 & 78.48 ± 0.17 & 85.63 ± 0.07 & 80.97 ± 5.29 & \underline{87.63 ± 0.52} & 86.06 ± 0.47 & 79.06 ± 0.38 & 86.63 ± 0.50 & \textbf{88.57 ± 0.37} \\
          & Ma-AP (↑) & \underline{24.65 ± 1.29} & 14.11 ± 0.12 &	9.65 ± 0.35 & 20.14 ± 0.13 & 13.70 ± 1.72	& 23.78 ± 1.17 & 20.94 ± 1.69 & 21.17 ± 0.61 & 20.60 ± 0.34	& \textbf{26.91 ± 1.67} \\
          & Mi-AP (↑) & 46.46 ± 2.06 & 35.11 ± 0.10 & 26.31 ± 0.34 & 36.61 ± 0.13 & 31.88 ± 7.37 & \underline{47.76 ± 1.59} & 42.10 ± 2.61 & 35.53 ± 0.18 & 41.30 ± 1.22 & \textbf{48.97 ± 1.30} \\
          & LRAP (↑) & 64.66 ± 0.80 & 55.45 ± 0.19 & 52.04 ± 0.13 & 59.71 ± 0.25 & 52.50 ± 3.97 & \underline{64.91 ± 1.14} & 61.02 ± 1.45 & 54.27 ± 0.38 & 62.46 ± 1.46 & \textbf{65.40 ± 0.83}\\
    \midrule
    \multirow{7}[4]{*}{\rotatebox{90}{PCG}} & Ranking (↓) & 27.50 ± 0.58 & 27.30 ± 0.04 & 27.97 ± 0.10 & 27.31 ± 0.16 & 28.55 ± 0.34 & \underline{27.11 ± 0.44} & 28.26 ± 0.51 & 28.15 ± 0.56 & 28.45 ± 0.53 & \textbf{25.97 ± 0.57} \\
          & Hamming (↓) & 13.22 ± 0.30 & 12.96 ± 0.00 & 13.20 ± 0.01 & 13.16 ± 0.01 & 12.59 ± 0.20 & \underline{12.56 ± 0.31} & 12.59 ± 0.18 & 16.28 ± 0.16 & 12.65 ± 0.14 & \textbf{12.41 ± 0.23} \\
\cmidrule{2-12}          & Ma-AUC (↑) & 62.92 ± 0.84 & 48.03 ± 0.36 & 45.60 ± 0.27 & 57.32 ± 0.41 & 53.84 ± 0.90 & 61.48 ± 0.84 & 58.75 ± 1.05 & \underline{63.02 ± 0.97} &  57.00 ± 1.14 & \textbf{64.86 ± 1.05} \\
          & Mi-AUC (↑) & 71.74 ± 0.38 & 67.61 ± 0.13 & 62.86 ± 0.13 & 70.89 ± 0.09 & 70.09 ± 0.38 & \underline{72.03 ± 0.40} & 71.31 ± 0.56 & 71.08 ± 0.62 & 70.02 ± 0.47 & \textbf{74.16 ± 0.61} \\
          & Ma-AP (↑) & \underline{23.49 ± 0.38} & 13.12 ± 0.05 & 12.68 ± 0.12 & 18.45 ± 0.29 & 16.01 ± 0.69 & 20.95 ± 0.91 & 19.49 ± 0.85 & 22.73 ± 0.75 & 16.23 ± 0.66 & \textbf{24.64 ± 0.90} \\
          & Mi-AP (↑) & \underline{30.04 ± 0.54} & 23.43 ± 0.09 & 21.32 ± 0.28 & 28.04 ± 0.29 & 25.50 ± 0.95 & 29.33 ± 1.21 & 27.54 ± 0.98 & 24.77 ± 0.85 & 24.87 ± 0.59 & \textbf{31.91 ± 1.07} \\
          & LRAP (↑) & 48.03 ± 1.25 & 46.42 ± 0.13 & 46.76 ± 0.24 & \underline{48.23 ± 0.26} & 45.37 ± 1.03 & 48.21 ± 0.57 & 46.62 ± 1.27 & 47.58 ± 0.86 & 45.76 ± 0.99 & \textbf{49.04 ± 0.83} \\
    \midrule
    \multirow{7}[4]{*}{\rotatebox{90}{Blogcatalog}} & Ranking (↓) & 25.67 ± 0.20 & 25.69 ± 0.02 & 25.95 ± 0.01 & \underline{25.48 ± 0.01} & 42.19 ± 3.87 & 25.68 ± 0.25 & \underline{25.48 ± 0.12} & 26.61 ± 0.86 & 25.67 ± 0.27 & \textbf{25.42 ± 0.23} \\
          & Hamming (↓) & 3.58 ± 0.03 & 3.57 ± 0.00 & \underline{3.56 ± 0.00} & 3.59 ± 0.00 & 3.58 ± 0.03 & 3.58 ± 0.03 & \textbf{3.55 ± 0.03} & 3.66 ± 0.04 & 3.58 ± 0.03 & 3.58 ± 0.03 \\
\cmidrule{2-12}          & Ma-AUC (↑) & 50.59 ± 0.51 & 48.19 ± 0.04 & 47.08 ± 0.01 & 50.52 ± 0.02 & 50.19 ± 2.06 & 50.94 ± 1.17 & \underline{53.29 ± 0.52} & 51.70 ± 0.32 & 50.52 ± 1.07 & \textbf{54.48 ± 0.52} \\
          & Mi-AUC (↑) & 73.80 ± 0.20 & 69.22 ± 0.05 & 65.28 ± 0.01 & 72.37 ± 0.01 & 56.74 ± 2.32 & 73.85 ± 0.17 & \underline{74.11 ± 0.06} & 71.75 ± 0.86 & 73.73 ± 0.47 & \textbf{74.15 ± 0.21} \\
          & Ma-AP (↑) & 4.16 ± 0.14 & 3.67 ± 0.01 & 3.67 ± 0.01 &	3.91 ± 0.02 & 4.10 ± 0.24 & 4.16 ± 0.14 & \textbf{5.07 ± 0.20} &	4.13 ± 0.13 & 4.15 ± 0.48 & \underline{4.61 ± 0.15} \\
          & Mi-AP (↑) & 9.41 ± 0.15 & 7.97 ± 0.01 & 7.13 ± 0.02 &	8.89 ± 0.02 & 4.72 ± 0.53 & 9.42 ± 0.15 & \textbf{11.71 ± 0.55} &	5.54 ± 1.33 & 9.38 ± 1.22 & \underline{9.65 ± 0.21} \\
          & LRAP (↑) & 27.88 ± 0.15 & 27.30 ± 0.00 & 27.82 ± 0.04 & 27.81 ± 0.00 & 17.58 ± 3.93 & 27.87 ± 0.18 & \textbf{28.94 ± 1.00} & 25.37 ± 0.27 & 28.10 ± 0.92 & \underline{28.32 ± 0.29} \\
    \midrule
    \multirow{7}[4]{*}{\rotatebox{90}{PPI}} & Ranking (↓) & \underline{18.32 ± 0.16} & 25.73 ± 0.01 & OOM & 25.44 ± 0.01 & 25.33 ± 0.08 & 20.05 ± 0.62 & 18.44 ± 0.14 & 18.38 ± 0.30 & 19.85 ± 0.18 &  \textbf{16.17 ± 0.22} \\
          & Hamming (↓) & 22.63 ± 0.14 & 26.28 ± 0.00 & OOM & 26.24 ± 0.00 & 25.60 ± 0.18 & 23.83 ± 0.14 & \underline{21.85 ± 0.28} & 31.06 ± 0.32 & 23.41 ± 0.13 & \textbf{20.79 ± 0.29} \\
\cmidrule{2-12}          & Ma-AUC (↑) & 73.06 ± 0.21 & 44.65 ± 0.02 & OOM & 51.82 ± 0.12 & 58.86 ± 0.29 & 70.26 ± 1.12 & \underline{74.24 ± 0.42} & 74.04 ± 0.41 & 70.21 ± 0.24 & \textbf{77.54 ± 0.42} \\
          & Mi-AUC (↑) & 80.19 ± 0.14 & 67.84 ± 0.01 & OOM & 70.24 ± 0.02 & 72.48 ± 0.08 & 78.37 ± 0.72 & \underline{81.15 ± 0.26} & 80.46 ± 0.26 & 78.50 ± 0.13 & \textbf{83.35 ± 0.29} \\
          & Ma-AP (↑) & 54.70 ± 0.42	& 29.34 ± 0.01	& OOM & 34.07 ± 0.01 & 39.06 ± 0.30 & 50.69 ± 1.54 & \underline{57.19 ± 0.15} &	55.71 ± 0.70 & 51.16 ± 0.61 & \textbf{61.33 ± 0.67} \\
          & Mi-AP (↑) & 67.64 ± 0.43 & 55.66 ± 0.01 & OOM & 56.52 ± 0.01 & 58.68 ± 0.37 & 64.98 ± 1.32 & \underline{69.59 ± 0.27} & 63.11 ± 0.62 & 65.57 ± 0.54 & \textbf{72.35 ± 0.58} \\
          & LRAP (↑) & 68.82 ± 0.39 & 62.19 ± 0.01 & OOM & 62.28 ± 0.01 & 62.18 ± 0.40 & 66.68 ± 1.01 & \underline{68.87 ± 0.35} & 63.68 ± 0.61 & 67.96 ± 0.36 & \textbf{71.40 ± 0.34} \\
    \midrule
    \multirow{7}[4]{*}{\rotatebox{90}{Delve}} & Ranking (↓) & 3.46 ± 0.04 & OOM & OOM & 6.10 ± 0.53 & OOM & 3.33 ± 0.01 & OOM & 5.07 ± 0.02 & \textbf{1.46 ± 0.02} & \underline{2.41 ± 0.02} \\
          & Hamming (↓) &  3.16 ± 0.01 & OOM & OOM & 30.26 ± 0.51 & OOM &  2.99 ± 0.01 & OOM & 7.50 ± 0.01 & \textbf{1.67 ± 0.07} & \underline{2.48 ± 0.01} \\
\cmidrule{2-12}          & Ma-AUC (↑) &  94.98 ± 0.14 & OOM  & OOM  & 58.37 ± 0.74 & OOM & 95.71 ± 0.04 & OOM & 95.26 ± 0.09 &  \textbf{97.85 ± 0.05} & \underline{96.84 ± 0.10} \\
          & Mi-AUC (↑) & 96.55 ± 0.06 & OOM & OOM & 57.58 ± 0.04 & OOM & 97.29 ± 0.02 & OOM & 95.11 ± 0.04 & \textbf{98.39 ± 0.07} & \underline{98.05 ± 0.02}\\
          & Ma-AP (↑) &  64.28 ± 0.39 & OOM & OOM & 11.60 ± 0.07  & OOM & 65.20 ± 0.07	& OOM & 65.67 ± 0.31 & \textbf{80.16 ± 0.32} & \underline{72.80 ± 0.26}\\
          & Mi-AP (↑) & 78.72 ± 0.07 & OOM & OOM & 10.12 ± 0.07 & OOM & 81.78 ± 0.09 & OOM & 68.97 ± 0.25 & \textbf{90.44 ± 0.12} & \underline{86.12 ± 0.03} \\
          & LRAP (↑) & 85.57 ± 0.10 & OOM & OOM & 29.44 ± 0.40 & OOM & 86.12 ± 0.05 & OOM & 76.47 ± 0.27 & \textbf{92.09 ± 0.21} & \underline{89.20 ± 0.01} \\
    \midrule
    \multicolumn{2}{c|}{\textit{Average Rank}} & \textit{3.86} & \textit{7.79} & \textit{8.52} & \textit{6.17} & \textit{7.79} & \textit{3.57} & \underline{\textit{3.50}} & \textit{5.94} & \textit{4.46} & \textbf{\textit{1.37}} \\   
    \bottomrule
    \end{tabular}%
    }
  \label{tab:main_results}%
\end{table*}%

\subsection{Main Results (RQ1)}
In this subsection, we compare our proposed \model with nine state-of-the-art baseline models on the five experimental datasets. The comparison results on seven metrics are reported in Table \ref{tab:main_results}, and corresponding Friedman statistics are shown in Table~\ref{tab:f_test}, with the following observations:
\begin{itemize}[leftmargin=*]
    \item \textbf{\model can achieve significant improvements over state-of-the-art methods on all experimental datasets.} From the table, we can observe that our proposed \model can generally achieve the best performance among the seven metrics on average. Specifically, \model outperforms the best baseline by 6.43\%, 2.92\%, 2.23\%, and 4.45\% for Macro-AUC on Humloc, PCG, Blogcatalog, and PPI, respectively.
    The significance of model gains was further validated by Friedman statistics and the Bonferroni-Dunn test. These results verify that designing the correlation decomposed graph and the equipped \model contributes to achieving better multi-label node classification performance.
    
    \item \textbf{Graph structure learning models do not bring about performance gains.}
    Comparing three graph structure learning models (GRCN, IDGL, SUMLIME) with the simply adopted GCN, we can find that these methods fail to yield substantial benefits in multi-label node classification scenarios and, in many instances, may even lead to negative outcomes.
    For example, all three models negatively impact the GCN model performance in terms of Micro-AUC, Macro-AP, and Micro-AP.
    This indicates that directly learning graph structures without considering multi-label correlations is less effective for the multi-label node classification downstream task.
    
    \item \textbf{The models with multi-label modeling generally become the best baselines.} We can observe from the comparison results that LANC and ML-GCN achieve the top-2 performance among the other baseline models. Their shared advantage lies in the explicit modeling of multi-label relationships. However, due to the unified message passing with ambitious information, they are less optimal compared to the \model. Our proposed \model explicitly considers the label correlation and addresses the ambitious message passing with enhanced information aggregation over the correlation-aware decomposed graph, which achieves the best performance.
\end{itemize}

We have also assessed the convergence of \model, with the results detailed in Appendix~\ref{sec:converge}. The results indicate that our model adeptly balances effectiveness and efficiency, also exhibiting faster convergence compared to the pure backbone. 

\begin{table}[tbp]
  \centering
  \caption{Friedman statistics $F_{F}$ on each evaluation metric with corresponding $p$-value (all < 0.05 level).}
    \resizebox{0.6\linewidth}{!}{
    \begin{tabular}{c|cc}
    \toprule
    Evaluation Metric & $F_{F}$  & $p$-value \\
    \midrule
    Ranking Loss & 17.59 & 0.0015 \\
    Hamming Loss & 21.97    & 0.0002 \\
    Macro-AUC & 11.80 & 0.0189 \\
    Micro-AUC & 13.41    &  0.0094 \\
    Macro-AP &  14.85 & 0.0050 \\
    Micro-AP & 16.54  & 0.0024 \\
    Label Ranking AP  & 16.54  & 0.0024 \\
    \bottomrule
    \end{tabular}%
    }
  \label{tab:f_test}%
\end{table}%

\subsection{Ablation Study (RQ2)}

\begin{figure}[tbp]
     \centering
     \includegraphics[width=\linewidth, trim=0cm 0cm 0cm 0cm,clip]{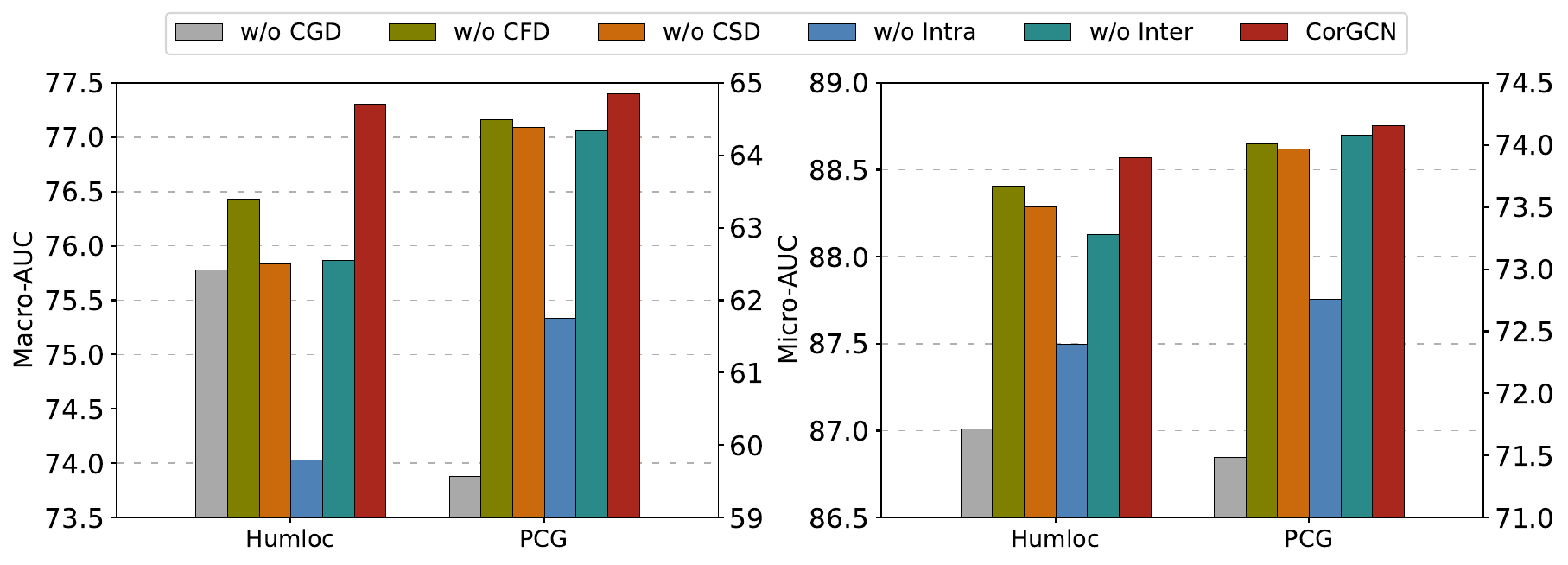}
     \caption{Abalation study on \model with its five variants.}
     \label{fig:ablation}
\end{figure}

To verify the effectiveness of the key components in \model, we conduct ablation studies by comparing \model with its five variants:
(1) \textbf{\textit{w/o CGD}} replaces the decomposed graphs by the origin input graph.
(2) \textbf{\textit{w/o CFD}} removes the correlation-aware decomposition of node features, replacing them with original features for graph learning.
(3) \textbf{\textit{w/o CSD}} excludes the correlation-aware structure decomposition, replacing them with correlated node features and origin structure for \model message passing.
(4) \textbf{\textit{w/o Intra}} ignores the intra-label message passing and merges all views of graphs for unified message passing.
(5) \textbf{\textit{w/o Inter}} ignore the inter-label correlation propagation without considering message correlations.
From Figure~\ref{fig:ablation}, we can have following observations:

\begin{itemize}[leftmargin=*]
    \item \textbf{Effectiveness of the correlation decomposed graph.} The severe performance decline of \textit{w/o CGD}, along with the diminished efficacy of \textit{w/o CFD} and \textit{w/o CSD}, demonstrates that the original graph topology and node features are suboptimal for multi-label node classification. The proposed correlation decomposed graph (CDG) with both correlation decomposed node features and topology structure can effectively benefit the classification.
    \item \textbf{Effectiveness of the correlation decomposed graph convolution.} Based on the effective correlation decomposed graph, the lack of carefully designed decomposed graph convolution modules (\textit{w/o Intra} and \textit{w/o Inter}) both results in inferior classification performance than \model, which demonstrates that modeling intra-label message and passing inter-label correlation is meaningful. Further, removing the intra-label message passing with unified message passing has a greater impact on \model, indicating that the issue of feature and topology ambitiousness indeed brings negative effects during the message passing.
\end{itemize}

\begin{figure}[tbp]
     \centering
     \includegraphics[width=\linewidth, trim=0cm 0cm 0cm 0cm,clip]{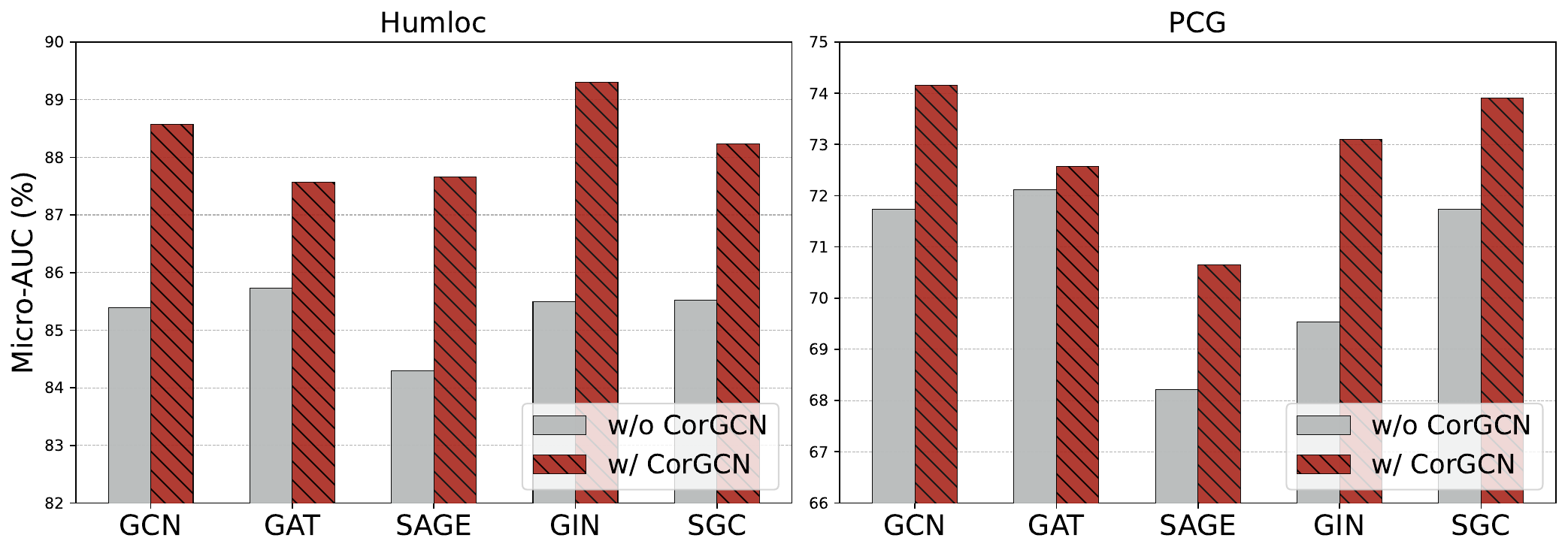}
     \caption{Generalization study results of \model in different GNN message passings.}
     \label{fig:backbone}
\end{figure}

\subsection{Generalization Study (RQ3)}
To investigate the generalization of the proposed \model method, we evaluate the performance of \model with different GNN backbones.
Specifically, we utilize the representative GCN-like~\cite{kipf2017gcn}, GAT-like~\cite{velivckovic2018gat}, SAGE-like~\cite{NIPS2017_graphsage}, GIN-like~\cite{xu2018gin}, and SGC-like~\cite{wu2019simplifying} for \model in Eq.(\ref{eq:gnn_mp}) of message passing, respectively.
The study results are illustrated in Figure~\ref{fig:backbone}. 

From the results, we can find that equipped with \model (w/ \model) is consistently significantly better than message passing with vanilla backbones (w/o \model), which brings 3.50\% and 3.15\% average Micro-AUC improvements on Humloc and PCG, respectively.
The improvements brought by \model to other backbones are greater compared to GAT. This is because GAT implicitly differentiates the neighborhoods by considering the importance of different nodes. However, the message passing paradigms in these backbones all remain ambitious in multi-label node classification. 
\model explicitly addresses this issue and thus exhibits better performance. 
The results demonstrate both the effectiveness and generalization of \model in different message passing functions.

\subsection{Case Study (RQ4)}\label{sec:case}
\begin{figure}[tbp]
     \centering
     \includegraphics[width=\linewidth, trim=0cm 0cm 0cm 0cm,clip]{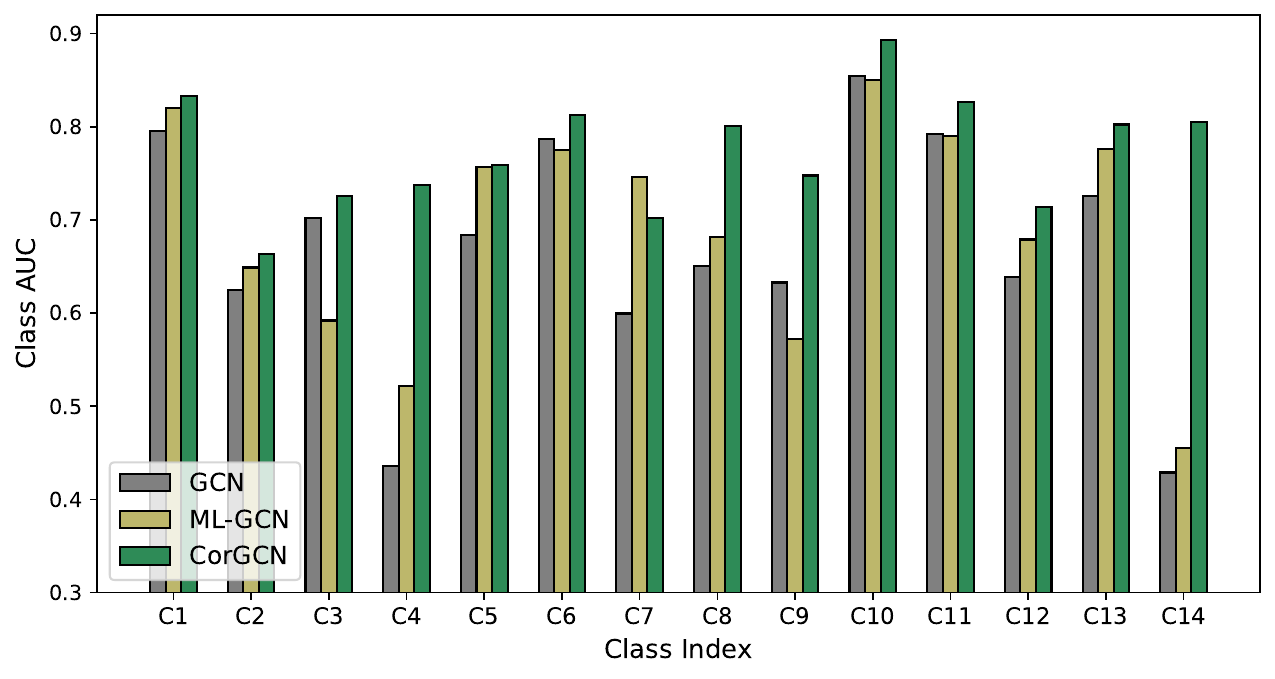}
     \caption{Case study of class AUC performance on Humloc.}
     \label{fig:case_study}
\end{figure}

To further analyze the performance of our CorGCN with the graph decomposition, we conduct a case study that analyzes the performance of each fine-grained class. The comparison between the original GCN and the top-performed baseline MLGCN over the Humloc dataset is as follows.
From the results in Figure~\ref{fig:case_study}, we can have the following findings: (i) Under the decomposition paradigm, CorGCN can achieve significantly better performance among all the fine-grained classes than the pure backbone. (ii) Although MLGCN is the top-performing baseline on Humloc, due to the lack of ambiguous discrimination ability with a unified convolution process, it will lead to 5/14 classes with performance dropping than the original GCN.

Therefore, based on the case study results, we can find that the decomposition paradigm of CorGCN is helpful in better understanding each class characteristic over the multi-label graph with ambiguous features and topology.

\subsection{Efficiency Study (RQ5)}\label{sec:effiency}
To further evaluate whether \model can balance the effectiveness and efficiency, we conduct the efficiency study of \model compared with the top-3 performed baselines on both the training and inference stages in Table~\ref{tab:efficiency}.

From the results, we can find that our \model can generally balance effectiveness and efficiency. For the training stage, the end-to-end label correlation modeling strategy employed by \model yields an efficiency that is second to that of the original GCN. The negative sampling process costs additional time for ML-GCN during training. Then, during the inference phase, \model has computational efficiency that is lower than top-3 performing baselines due to the multi-label structure decomposition, while demonstrating a good enhancement in performance compared to these models.

\begin{table}[htbp]
  \centering
  \caption{Efficiency study results on Humloc and PCG datasets each convolution epoch in seconds.}
  \resizebox{0.92\linewidth}{!}{
    \begin{tabular}{ccccc|c}
    \toprule
    Dataset & Stage & GCN   &  ML-GCN & LANC & \textbf{\model} \\
    \midrule
    \multirow{2}[1]{*}{{Humloc}} & Training & 0.37 &	15.79 &	1.18 & 0.76	\\
          & Inference & 0.37	& 0.49	& 0.48 & 0.57 \\   
    \midrule
    \multirow{2}[1]{*}{{PCG}} & Training & 0.39	& 18.15	& 1.21 & 0.84 \\
          & Inference 	& 0.30 &	0.62	& 0.39 & 0.64\\
    \bottomrule
    \end{tabular}%
    }
  \label{tab:efficiency}%
\end{table}%

\subsection{Parameter Study (RQ6)}
\subsubsection{Effect of Parameter $\lambda$}
We investigate the effect of $\lambda$ for density controlling in correlation decomposed structure learning with the range from 1 to 13 with a step size of 2 as Figure \ref{fig:param}.
From the results, we can observe that a too-small value of $\lambda$ will cause poor performance, which indicates the information is too sparse and insufficient. Furthermore, the suitable value of $\lambda$ for Humloc is larger than the value for PCG, one possible reason is that the $\lambda$ of \model should be set larger on more sparse graphs.

\subsubsection{Effect of Macro Label Prototype Number $K'$}
\begin{figure}[tbp]
     \centering
     \includegraphics[width=\linewidth, trim=0cm 0cm 0cm 0cm,clip]{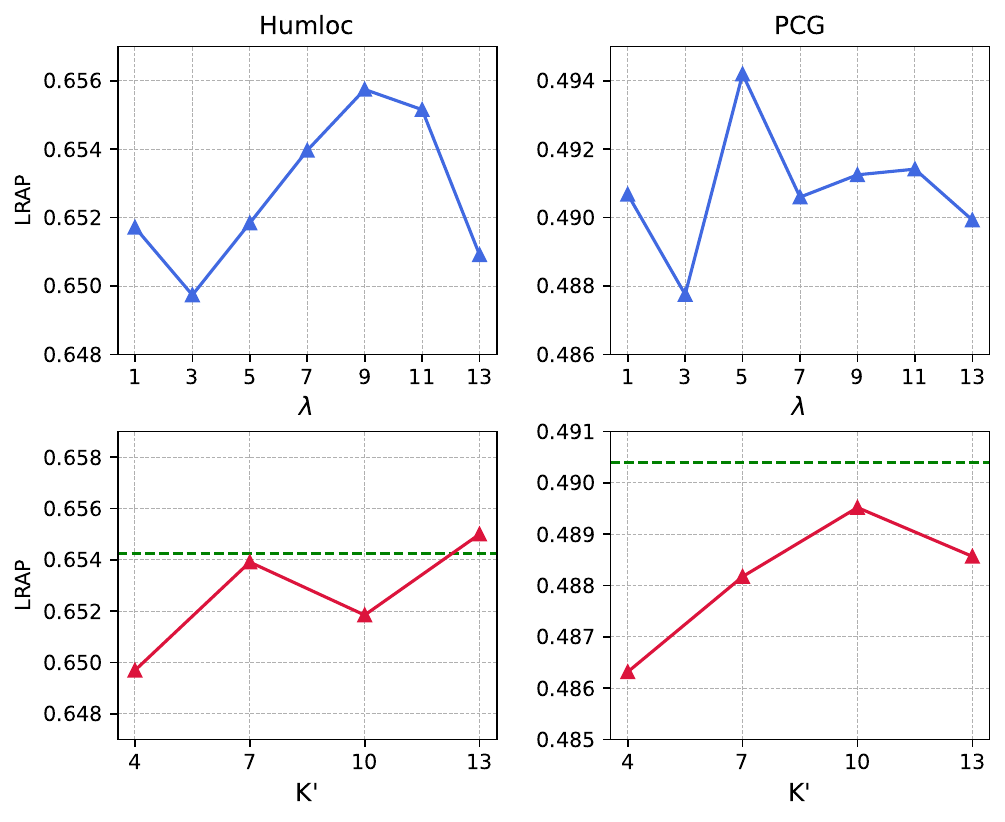}
     \caption{Parameter study results of $\lambda$ and $K'$.}
     \label{fig:param}
\end{figure}

We also evaluate the impact of different macro label numbers $K'$ in Sec.~\ref{sec:extend_macro} to \model.
From Figure \ref{fig:param}, we can find that the too-small cluster number will lead to too coarsen label segmentation and result in relatively poor results.
In addition, choosing a relatively appropriate number of macro labels, such as 13 for Humloc and 10 for PCG, can achieve model performance close to or even equivalent to the performance of original label numbers (the green dashed line).
More hyperparameter studies on $\gamma$ for \model can be found in Appendix~\ref{sec:add_param}.

\section{Conclusion}
In this paper, we propose Correlation-Aware Graph Convolutional Networks (\model) for multi-label node classification. Specifically, aiming to address the ambitious feature and topology in the original graph in multi-label scenarios, we introduce more suitable Correlation-aware Decomposed Graphs (CDG). Then, based on the CDG, we further design a novel Correlation-enhanced Graph Convolution with inter-label message passing and intra-label correlation propagation. Extensive experiments on five multi-label node classification datasets and in-depth analyses in multiple perspectives illustrate the effectiveness and strengths of \model. Future works can be conducted on the optimization of model inference efficiency and the exploration of downstream tasks of multi-label node classification, such as multi-label graph anomaly detection.

\begin{acks}
This work is supported by the National Natural Science Foundation of China (No. 62476245, No. 62372408), Zhejiang Provincial Natural Science Foundation of China (Grant No. LTGG23F030005). Carl Yang was not supported by any grants from China.
\end{acks}

\bibliographystyle{ACM-Reference-Format}
\bibliography{reference}

\appendix
\section{Methodology Details}\label{sec:add_analysis}
\subsection{Analysis of the Objective Function on the Multi-Label Estimator}\label{sec:add_mie}

The mutual information estimator $\phi$ and likelihood maximizing decoder $\theta$ work collaboratively for node-label correlation and label-label correlation modeling as
\begin{equation}
    \mathop{\arg\max}\limits_{\phi, \theta}\underbrace{I_{\phi}(\bm{E}^{x}, \bm{E}^{l})}_{\text{Mutual Information}} + \underbrace{\mathcal{L}_{\theta}(\theta; \bm{E}^{x}) + \mathcal{L}_{\theta}(\theta; \bm{E}^{l})}_{\text{Likelihood}}.
\end{equation}
Analysis of the effectiveness of the mutual information estimator $\phi$ for multi-label correlation modeling can be found in previous works~\cite{bai2022gaussian,zhao2021hot}.
Then, the likelihood maximizing decoder $\theta$ can be utilized as a likelihood estimator of node feature and label prototype with the entropy KL divergence with focal enhancement~\cite{lin2017focal} for target labels as follows:
\begin{equation}\label{eq:leapp}
\begin{aligned}
    \mathcal{L}_{lm} = \frac{1}{2|\gV_L|} \sum_{v_i \in \gV_L} \sum_{k=1}^{K} & \rho_{k}[\underbrace{(1 - p_{k}(\bm{E}_{i}^x | \theta))^{\gamma}\text{log}p_{k}(\bm{E}_{i}^x | \theta)}_{\text{Likelihood Estimation with Node Feature}} \\
    &+ \underbrace{(1 - p_{k}(\bm{y}_{i}\cdot \bm{E}^{l} | \theta))^{\gamma}\text{log}p_{k}(\bm{y}_{i}\cdot \bm{E}^{l} | \theta)}_{\text{Likelihood Estimation with Label Prototype}}].
\end{aligned}
\end{equation}

\vspace{-1em}
\subsection{Complexity Analysis}~\label{sec:complex_analyse}
During the correlation-aware graph decomposition, the calculation involves the projection process, incurring a computational cost of \(\Theta(K \cdot n^2)\). 
For the Correlation-Enhanced graph convolution, we postulate that the message passing process within a graph comprising \(n\) nodes entails a cost of \(\Theta_{\text{mp}}(n)\), which varies according to the specific GCN backbone employed. During the intra-label message passing, since the process can be parallelized across $K$ label views, the overall expense of message passing is quantified as \(\Theta_{\text{mp}}(n)\).
Then, the cost associated with inter-label correlation propagation is \(\Theta(K^2\cdot n)\).
Thus, the overall time complexity of the process can be expressed as \(\mathbf{T}(n) = \Theta(K(n^2 + \Theta_{\text{mp}}(n) + K\cdot n))\). 

As outlined in Sec.~\ref{sec:extend_macro}, this complexity can be reduced to \(\mathbf{T}(n) = \Theta(K'(n^2 + \Theta_{\text{mp}}(n) + K'n))\) for the large label space, where \(K' \ll K\), denoting a significant reduction in computational cost.
In practice, the entire process can be batch-processed, where the similarity is calculated among the nodes within each batch, thus making the time complexity of each batch $B$ to \(\mathbf{T}(B) = \Theta(K'(B^2 + \Theta_{\text{mp}}(B) + K'B))\). Computational operations of $K'$ labels can be performed in parallel by constructing a matrix tensor for $K'$ label spaces.

\vspace{-0.5em}
\subsection{Pseudocode of \model}\label{sec:model_alg}
\model includes (1) correlation-aware graph learning with feature and topology decompositions, and (2) correlation-aware graph convolution with intra-label message passing and inter-label correlation propagation. The general process is described in Algorithm~\ref{alg:framework}.

\begin{algorithm}[tbp]
    \renewcommand{\algorithmicrequire}{\textbf{Input:}}  
    \renewcommand{\algorithmicensure}{\textbf{Output:}} 
    \caption{The overall workflow of \model.}  
    \label{alg:framework}
    \begin{algorithmic}[1]
        \REQUIRE Feature matrix $\bm{X}$; Label prototype $\bm{E}^l$; Graph topology $\bm{A}^{0}$; Number of GNN layers $N_{gnn}$.
        \ENSURE Predicted labels $\bm{\hat{y}}$.
        \STATE $E^x \leftarrow \bm{X}$; //Compute transformed features
        \STATE $\mathcal{L}_{cmi} \leftarrow \{\bm{E}^x,\bm{E}^l\}$ using Eq.(4); // Compute contrastive loss
        \STATE $\mathcal{L}_{lm} \leftarrow \{\bm{E}^x,\bm{E}^l,\bm{y}\}$ using Eq.(5) and Eq.(6); // Compute likelihood loss
        \STATE $\bm{E}^{proj} \leftarrow \{\bm{E}^x,\bm{E}^l\}$ using Eq.(7), (8); // Compute projected features as node representations
        \STATE $\bm{E}^{sd} \leftarrow \{\bm{E}^{proj},\bm{A}\}$ using Eq.(9); // Aggregate the node representations from neighborhood
        \STATE $CDG \leftarrow \{\bm{E}^{sd},\bm{E}^{proj},\bm{A}\}$ using Eq.(10) - Eq.(13); // Learn the graphs by aggregated representations
        \FOR{$epoch \in 1,2,...N_{gnn}$}
        \STATE $\bm{\hat{Z}}^{(l)} \leftarrow \{\bm{Z}^{(l-1)},CDG\}$ using Eq.(14); // Update the node representations by message passing
        \STATE $\bm{Z}^{(l)} \leftarrow \{\bm{\hat{Z}}^{(l)},\bm{E}^l\}$ using Eq.(15) - (16); // Update the node representations by correlation propagation
        \ENDFOR
        \STATE $\bm{\hat{y}} \leftarrow \{\bm{Z}^{(E)},\bm{E}^l\}$ using Eq.(17) - (18); // Predict the probabilities 
        \STATE $\mathcal{L}_{cls} \leftarrow \{\bm{\hat{y}},\bm{y}\}$ using Eq.(19); // Compute the classification loss
        \STATE $\mathcal{L} \leftarrow \mathcal{L}_{cls}+\alpha\mathcal{L}_{cmi}+\beta\mathcal{L}_{lm}$ ;
        \IF{$Training$}
        \STATE Back-propagate $\mathcal{L}$ to update model parameters.
        \ENDIF
    \end{algorithmic}  
\end{algorithm}

\section{Experimental details}
\subsection{Dataset Details}\label{sec:data_appendix}
We adopt five publicly available datasets for comprehensive evaluation of our proposed \model. The detailed description is as follows: 
\textbf{Humloc}\footnote{\url{https://github.com/Tianqi-py/MLGNC}\label{data_addr1}}~\cite{zhao2023multi} 
is a human protein subcellular location prediction dataset, consisting of 3,106 nodes and 18,496 edges.
Each node may have one or more labels in 14 possible locations. The edge data is derived from protein-protein interactions obtained from the open-sourced database. An edge exists between two nodes in the graph if there is an interaction between the respective proteins.
\textbf{PCG}\textsuperscript{\ref{data_addr1}}~\cite{zhao2023multi} is a protein phenotype prediction dataset with 3,233 nodes into 15 classes and 37,351 edges between them.
Each node represents a protein with 32-dimensional features. Each edge between a pair of protein nodes is their functional interaction.
The multi-label that each node is associated with is its correlated phenotypes, of which a phenotype is any observable characteristic or trait of a disease. The correspondence edge between protein and phenotype is retrieved from the open-sourced database.
\textbf{Blogcatalog}\footnote{\url{https://figshare.com/articles/dataset/BlogCatalog_dataset/11923611}}~\cite{zhou2021lanc} is a social network dataset with 10,312 nodes and 333,983 edges, which each node represents a blogger at the platform and each edge represents the contact relationship between a pair of nodes. The labels denote the social interest groups a blogger node is a part of, forming a total of 39 interest groups.
\textbf{PPI}\footnote{\url{https://github.com/GraphSAINT/GraphSAINT}}~\cite{zeng2019graphsaint} serves as a dataset with a large label space for protein-protein interaction networks, comprising 14,755 nodes (representing proteins) and 225,270 edges (indicating functional interactions). The dataset is annotated with gene ontology categories as labels, encompassing a total of 121 unique sets.
\textbf{Delve}\footnote{\url{https://www.dropbox.com/sh/tg8nclx5gpctbmo/AADQc0YFpuVeqXhaNuUK4MMba?dl=0}}~\cite{xiao2022larn} is a large-scale citation network dataset with 1,229,280 nodes and 4,322,275 edges, which each node is a paper and each edge represents a citation relationship between a pair of papers. The labels denote the research fields that papers belong to, forming a total of 20 categories.

\subsection{Baseline Details}\label{sec:model_appendix}
We compare our proposed \model with nine representative state-of-the-art GNN models as follows:
\textbf{GCN}~\cite{kipf2017gcn} is a widely recognized model in the domain of graph neural networks, grounded in spectral theory.
\textbf{GRCN}~\cite{yu2021grcn} incorporates a module based on GCN specifically engineered for the prediction of missing edges and the adjustment of edge weights, which is optimized with downstream tasks.
\textbf{IDGL}~\cite{chen2020idgl} mutually enhances the graph structure and node embeddings: the premise is to iteratively improve the graph structure through enhanced node embeddings and, reciprocally, to refine node embeddings leveraging an optimized graph structure.
\textbf{SUBLIME}~\cite{liu2022sublime} uses a contrastive loss to maximize the agreement between the anchor graph and the learned graph, which is further enhanced with a bootstrapping mechanism.
\textbf{GCN-LPA}~\cite{wang2021gcnlpa} unifies GCNs and label propagation in the same learning framework, which we utilized to propagate the multi-label simultaneously.
\textbf{ML-GCN}~\cite{gao2019mlgcn} employs a strategy to generate and utilize a label matrix for capturing node-label and label-label correlations through a relaxed skip-gram model in a unified vector space.
\textbf{LANC}~\cite{zhou2021lanc} develops a label attention module that employs an additive attention mechanism to synergize input and output contextual representations.
\textbf{SMLG}~\cite{song2021semi} comprises a joint variational representation module that generates node and label embeddings and a confidence-rated margin ranking module that detects second-order label correlations and refines the decision boundaries.
\textbf{LARN}~\cite{xiao2022larn} designs a novel label correlation scanner that adaptively captures label relationships, extracting vital encoded information for the generation of comprehensive representations.

\subsection{Implementation Details}\label{sec:imp_appendix}

We explore the combination space of the following hyper-parameters $lr$, $\lambda$, $K'$, $\gamma$ and list the values finally chosen in Table~\ref{tab:best_param} as the setting of performing the main results. Specifically, the learning rate $lr$ is selected from \{0.0001, 0.0005, 0.001, 0.005, 0.01, 0.02, 0.05\}; $\lambda$ that controls the density of graphs is tuned from 1 to 20; the macro label prototype number $K'$ is searched from 1 to 20; and $\gamma$ that controls the focusing factor in $\mathcal{L}_{lm}$ is chosen from \{1.0, 1.5, 2.0, 2.5, 3.0\}. Besides, to compare fairly, the hidden dim $d$ is set to 64, and the dropout rate of the network is set to 0.3. To balance the proportions between $\mathcal{L}_{cls}$, $\mathcal{L}_{cmi}$ and $\mathcal{L}_{lm}$, we calculate the loss balancing parameters $\alpha=|\frac{\mathcal{L}_{cls}}{3\mathcal{L}_{cmi}}|$ and $\beta=|\frac{\mathcal{L}_{cls}}{3\mathcal{L}_{lm}}|$ adaptively.

\begin{table}[htbp]
  \centering
  \small
  \caption{The values of hyper-parameters used in \model.}
    \begin{tabular}{c|c|c|c|c}
    \toprule
    Dataset & $lr$ & $\lambda$ & $K'$ & $\gamma$  \\
    \midrule
    Humloc & 0.001 & 7 & $\backslash$ & 2.0  \\
    PCG & 0.001 & 19 & $\backslash$ & 2.0 \\
    Blogcatalog & 0.001 & 5 & $\backslash$ & 2.0  \\
    PPI & 0.02 & 5 & 20 & 2.0 \\
    Delve &  0.001 & 5 & 10 &  2.0 \\
    \bottomrule
    \end{tabular}%
  \label{tab:best_param}%
\end{table}%

\section{Convergence Analysis}\label{sec:converge}
\begin{figure}[htbp]
\centering
  \subfigure[Training (Humloc)]{
    \centering
    \includegraphics[width=0.477\linewidth]{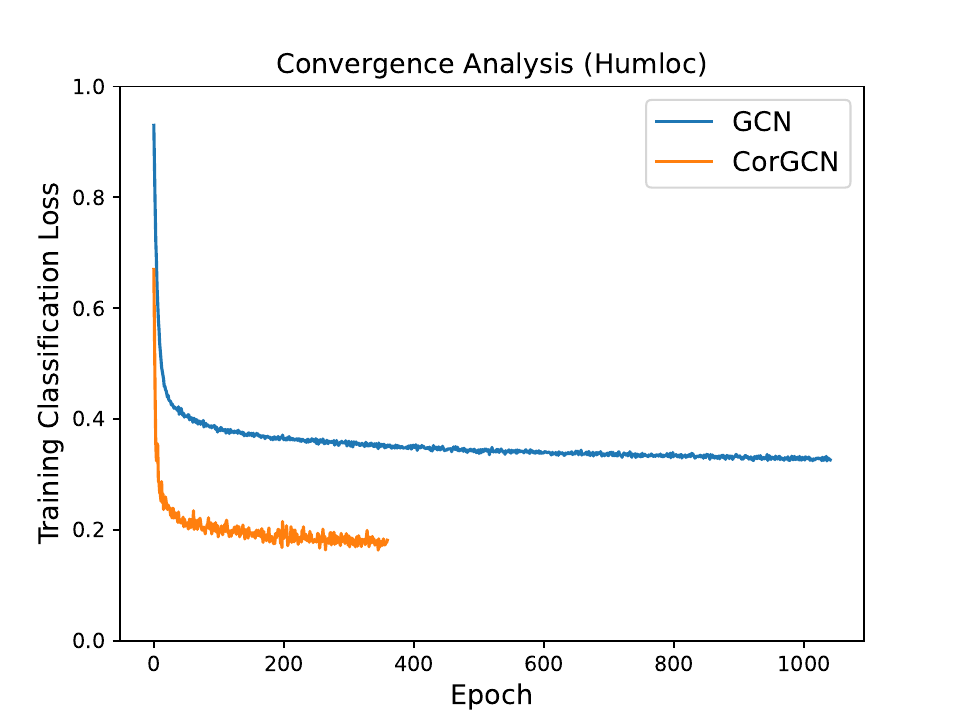}
  }
  \centering
  \subfigure[Training (PCG)]{
    \centering
    \includegraphics[width=0.477\linewidth]{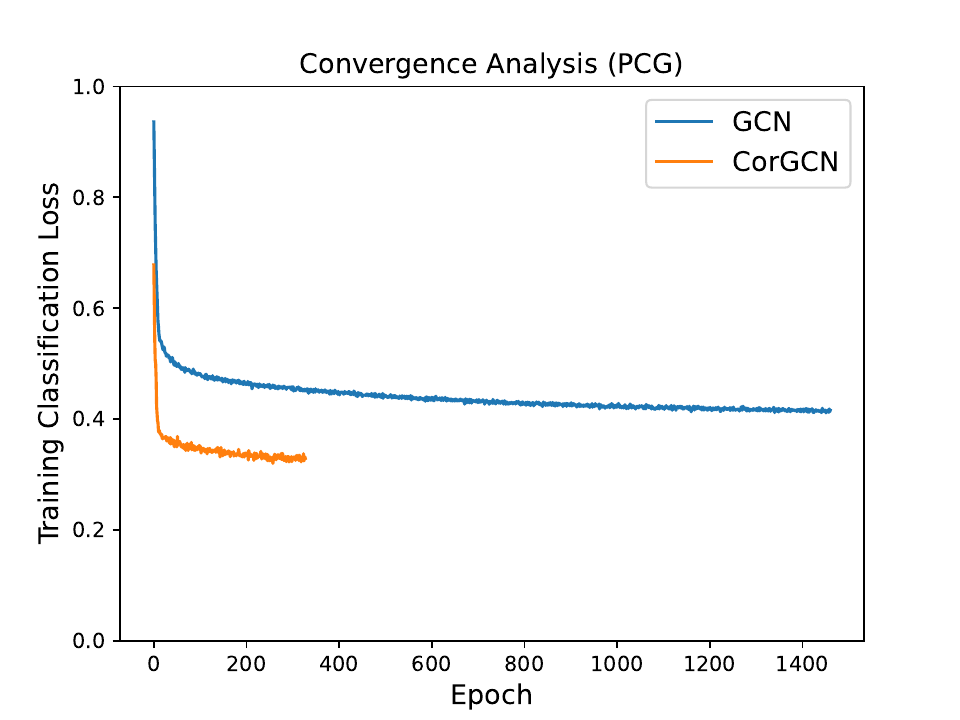}
  }
  \centering
  \subfigure[Validation (Humloc)]{
    \centering
    \includegraphics[width=0.477\linewidth]{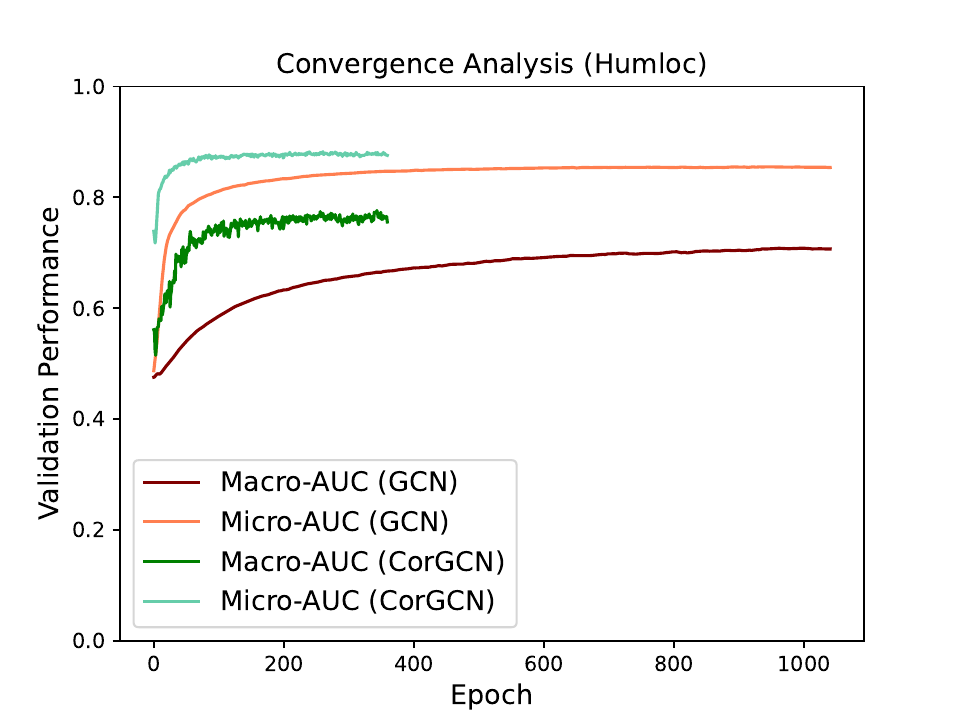}
  }
  \subfigure[Validation (PCG)]{
    \centering
    \includegraphics[width=0.477\linewidth]{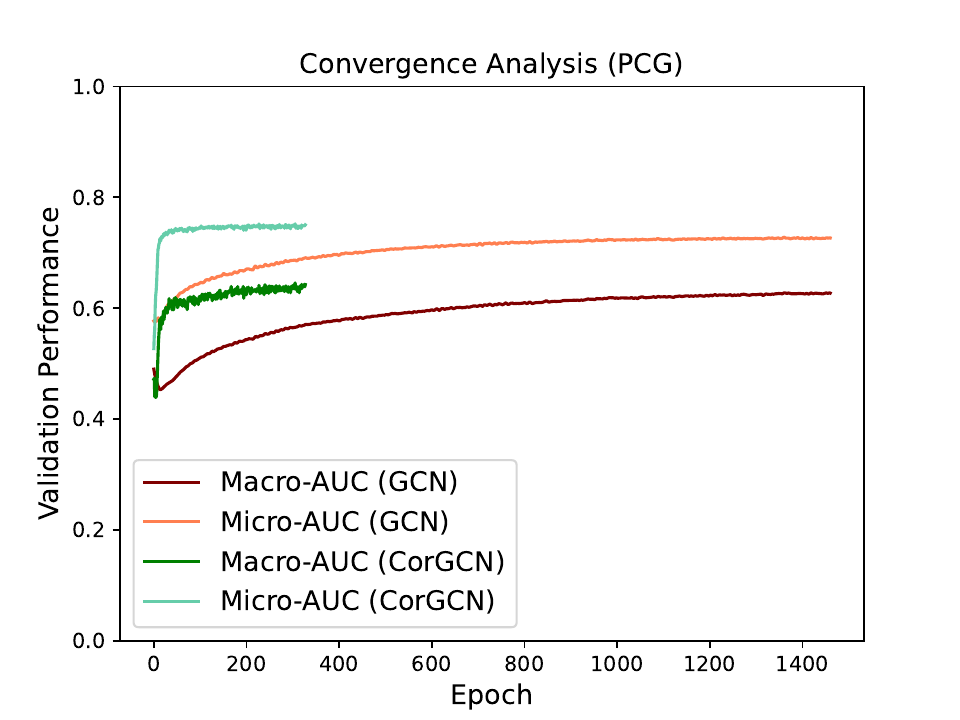}
  }
  \caption{Convergence analysis on (a)-(b) the training classification loss. (c)-(d) the validation set performance.}
  \vspace{-0.7em}
  \label{fig:converge}
\end{figure}

To further analyze the convergence of our CorGCN, we provide the figures of training classification loss changing over epochs and validation metrics changing over epochs as Figure~\ref{fig:converge}. From the results, we can find that \model can achieve a faster and better convergence than the pure backbone for multi-label node classification with lower training loss and higher validation performance. This is due to the design of \model with the correlation-aware multi-label graph decomposition and convolution can help the backbone better and faster understand the multi-label relationships and important information during message passing in the training process.

\section{More Parameter Study Results}\label{sec:add_param}

\begin{figure}[bp]
     \centering
      \vspace{-0.5em}
     \includegraphics[width=0.863\linewidth, trim=0cm 0cm 0cm 0cm,clip]{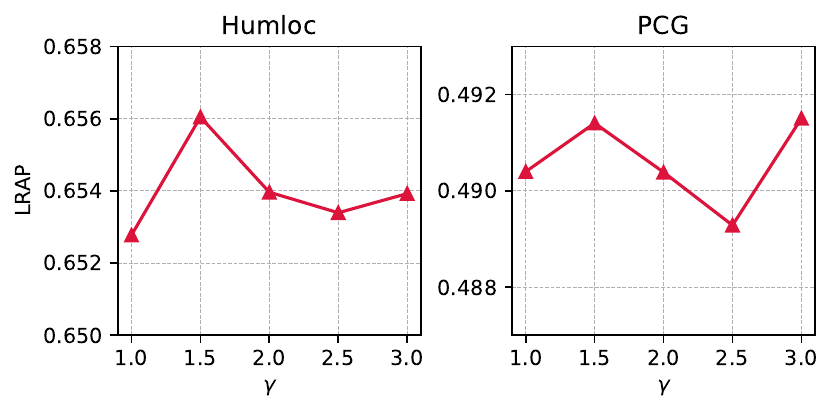}
     \vspace{-0.8em}
     \caption{Parameter study results on $\gamma$ of \model.}
     \label{fig:param_gamma}
\end{figure}

We explored the impact of $\gamma$ here. Parameter $\gamma$ amplifies the loss weight of misclassified samples, thereby increasing the model's attention to them. As shown in Figure~\ref{fig:param_gamma}, it can be observed that the model achieves the best performance on the LARP metric when $\gamma$ is set to 1.5. However, as $\gamma$ further increases, it may cause the model to overly focus on misclassified samples, neglecting the well-classified samples, resulting in a slight decrease in model performance.

\end{document}